\documentclass[10pt,twocolumn,letterpaper]{article}

\usepackage{cvpr}              
\usepackage{url}
\usepackage{tcolorbox}

\usepackage{booktabs}     
\usepackage{multirow}     
\usepackage{graphicx}     

\usepackage{algorithm}  
\usepackage{algpseudocode}  
\usepackage{amsmath}  
\usepackage{amssymb}  
\usepackage{multicol}
\usepackage{subcaption}

\usepackage{xspace}
\usepackage{pifont}
\usepackage[accsupp]{axessibility}

\renewcommand{\cite}{\citep}
\definecolor{cvprblue}{rgb}{0.21,0.49,0.74}
\usepackage[pagebackref,breaklinks,colorlinks,allcolors=cvprblue]{hyperref}

\def\paperID{42024} 
\def\confName{CVPR}
\def\confYear{2026}

\title{Unified Number-Free Text-to-Motion Generation Via Flow Matching}

\author{Guanhe Huang\\
King’s College London\\
{\tt\small guanhe.huang@kcl.ac.uk}
\and
Oya Celiktutan\\
King’s College London\\
{\tt\small oya.celiktutan@kcl.ac.uk}
}

\begin{document}
\maketitle

\begin{abstract}
    Generative models excel at motion synthesis for a fixed number of agents but struggle to generalize with variable agents. Based on limited, domain-specific data, existing methods employ autoregressive models to generate motion recursively, which suffer from inefficiency and error accumulation. We propose \textbf{Unified Motion Flow (UMF)}, which consists of Pyramid Motion Flow (P-Flow) and Semi-Noise Motion Flow (S-Flow).
    UMF decomposes the number-free motion generation into a single-pass motion prior generation stage and multi-pass reaction generation stages. Specifically, UMF utilizes a unified latent space to bridge the distribution gap between heterogeneous motion datasets, enabling effective unified training. 
    For motion prior generation, P-Flow operates on hierarchical resolutions conditioned on different noise levels, thereby mitigating computational overheads.
    For reaction generation, S-Flow learns a joint probabilistic path that adaptively performs reaction transformation and context reconstruction, alleviating error accumulation.
    Extensive results and user studies demonstrate UMF’s effectiveness as a generalist model for multi-person motion generation from text. 
    Project page: \url{https://githubhgh.github.io/umf/}.
\end{abstract}

\section{Introduction}


%
Text-to-motion generation, particularly via diffusion models, has advanced rapidly, progressing from single-agent~\cite{tevet2022human, guo2024momask, guo2022generating, guo2025motion, wang2025difusion} to multi-agent~\cite{liang2024intergen, xu2024inter, fan2025go, ruiz2025mixermdm, wu2025text2interact, shan2024towards} synthesis.
%
However, how to synthesize realistic number-free (\ie, any arbitrary number) human motions with text prompts remains an open challenge.
Existing methods struggle to generalize to unseen crowded scenes and are limited by motion data scarcity.
%
These limitations hinder the applications in robotics~\cite{liu2025takes, ji2025towards} and virtual reality~\cite{xu2025perceiving, Hong_2025_CVPR}, which often require seamless transitions between independent and collaborative tasks. 
This gap highlights the need for methods that can effectively utilize available heterogeneous data~\cite{fan2025freemotion, shafir2023human}.

%


\begin{figure}[t] 
\centering    
\includegraphics[width=0.45\textwidth]{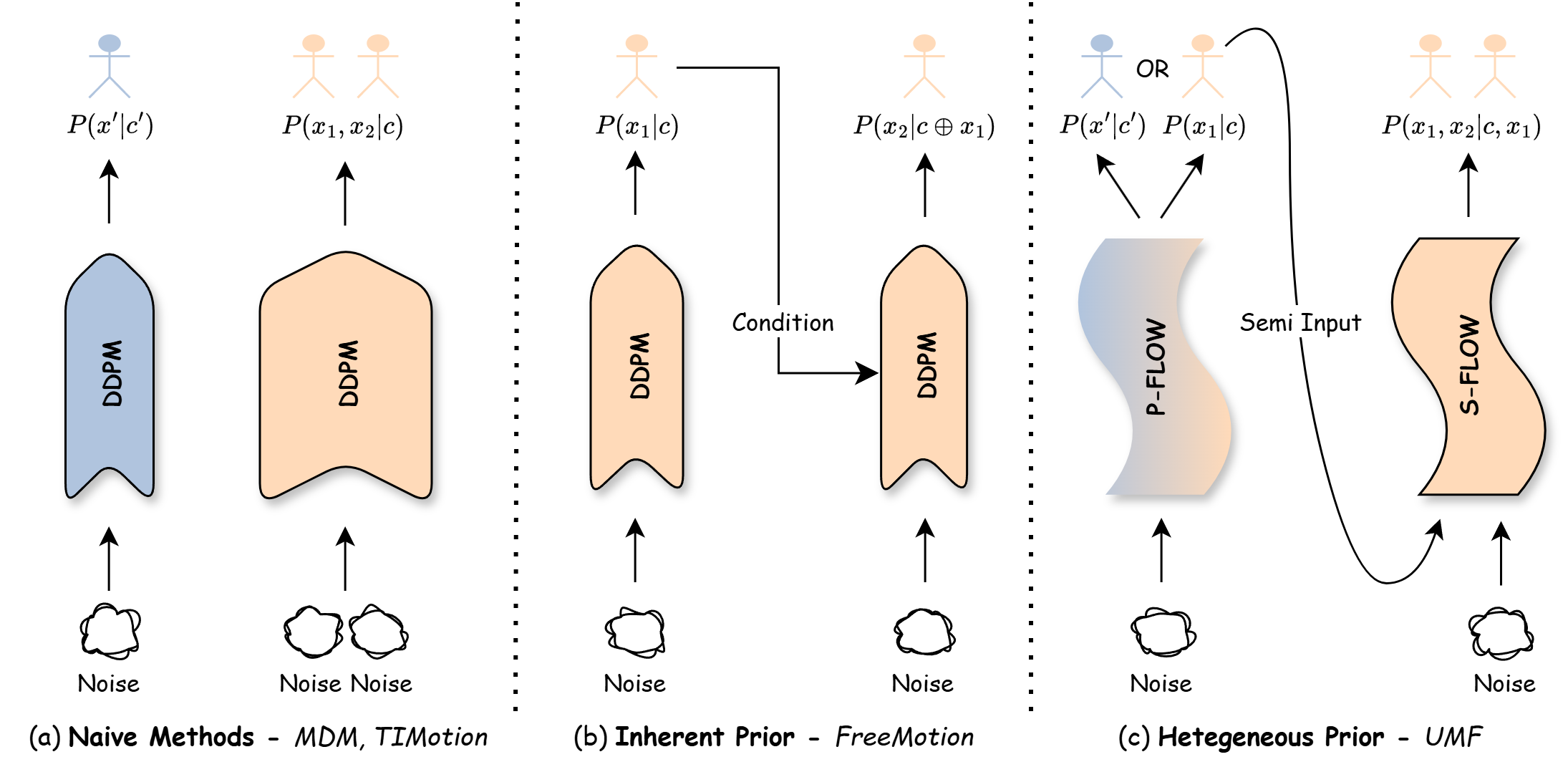}
\caption{
%
Core contribution of UMF. 
We show dual-agent cases here for simplicity. 
(a) Standard methods~\cite{tevet2022human, wang2025timotion} are restricted to a fixed number of agents. 
(b) Autoregressive methods~\cite{fan2025freemotion} decouple generation into a motion prior and subsequent reaction. The reaction is typically guided by the prior using a conditioning network.
(c) Our UMF leverages a heterogeneous motion prior as the adaptive start point of the reaction flow path, mitigating error accumulation.
}
\vspace{-5mm}
\label{intro_vis}
\end{figure}

%
To address the problem of text-to-motion generation with varying number of agents, previous methods typically rely on tailored architectures, more specifically, requiring expensive and time-consuming datasets~\cite{guo2022generating, liang2024intergen} for specific motion generation tasks.
Critically, existing multi-person interaction datasets~\citep{liang2024intergen, xu2024inter} are smaller and less diverse compared to single-person datasets~\cite{guo2022generating, mahmood2019amass, ionescu2013human3}, despite the interactive tasks being more complex.
On the other hand, there is a significant overlap for basic movements (\eg, walking) across these heterogeneous datasets, suggesting that \textit{single-person motion data can serve as heterogeneous prior for interaction synthesis}.

To leverage this overlap, in this paper, we introduce a single-person multi-token tokenizer that supports unified modeling and establishes the foundation for number-free, text-conditional generation.
Compared to the noisy raw motion space, the regularized multi-token latent space stabilizes flow matching training on heterogeneous single-agent (\ie, HumanML3D~\cite{guo2022generating}) and multi-agent (\ie, InterHuman~\cite{liang2024intergen}) datasets.
Based on this latent space, we propose Unified Motion Flow (UMF), a framework for number-free human motion generation from text prompts. UMF features two modules, the Pyramid Motion Flow (P-Flow) and Semi-Noise Motion Flow (S-Flow), which utilize flow matching to learn the mapping between text, motion prior, and reaction.
%
%
Specifically, it decouples number-free generation into a single-pass motion prior initialization (P-Flow) and a subsequent multi-pass reaction transformation (S-Flow).




Compared to previous single-token methods~\cite{chen2023executing, dai2024motionlcm}, our multi-token latent space shows superior reconstruction performance, mitigating heterogeneous domain gaps. However, it also imposes greater computational overhead.
Inspired by the fact that samples in early timesteps are noisy and less informative~\cite{wang2025lavie, jin2024pyramidal}, we introduce the P-Flow, which decomposes the motion prior generation into continuous hierarchical stages based on the timestep (noise level).  
Specifically, P-Flow maintains the original resolution only at later timesteps and applies a lower resolution via downsampling for early stages.
Previous works~\cite{teng2023relay, xie2024towards, jin2023act} that employ cascade models for these different resolutions are still accompanied by extra model complexity.
In contrast, our P-Flow can handle different resolutions within a single transformer~\cite{vaswani2017attention}, improving efficiency for multi-token motion prior generation.


The motion prior generated by P-Flow serves as the input for the iterative synthesis of subsequent agent reactions. However, this autoregressive process often suffers from potential error accumulation~\cite{ji2025denoising, wang2025erroranalysesautoregressivevideo}.
%
Previous methods~\cite{fan2025freemotion} rely on deterministic condition mechanisms (\eg, ControlNet~\cite{zhang2023adding}) to guide the process, which struggle to capture the causal relationship between interactive agents.
%
%
Consequently, we propose Semi-Noise Motion Flow (S-Flow) to learn the joint probabilistic path between previously generated motions (the context) and the subsequent agent's motion (the reaction).
%
As shown in Fig.~\ref{intro_vis}, rather than using the generated motions as a static condition, S-Flow integrates them to define the context distribution.
%
This source distribution initializes the reaction generation path, which enables S-Flow to focus directly on learning the dynamic transformation between motion distributions.
%
Concurrently, the S-Flow learns another auxiliary path to reconstruct the integrated context from noise distributions, as a strong regularizer for global interactive dependencies.
This joint training of two distinct flow paths balances between the reaction prediction and context awareness, making it less prone to error accumulation.

%
%
%
%


In summary, our contributions are as follows:
\begin{itemize}[itemsep=0pt, topsep=3pt, leftmargin=5pt]

\item We propose Unified Motion Flow (UMF), a generalist framework for number-free text-to-motion generation. UMF’s core design unifies heterogeneous single-person (e.g., HumanML3D) and multi-person (e.g., InterHuman) datasets within a multi-token latent space.

\item For efficient individual motion synthesis, we introduce Pyramid Motion Flow (P-Flow). P-Flow operates on hierarchical resolutions conditioned on the noise level, which alleviates computational overheads of multi-token representations while maintaining high-fidelity generation.

\item For reaction and interaction synthesis, we develop Semi-Noise Motion Flow (S-Flow). S-Flow learns a joint probabilistic path by balancing reaction transformation and context reconstruction, thereby alleviating error accumulation.

\item Extensive experiments demonstrate UMF achieves state-of-the-art (SoTA) performance for multi-person generation (FID 4.772 on InterHuman) benchmarks. We also validate UMF's zero-shot generalization to unseen group scenarios through a user study.
%
\end{itemize}





\section{Related Work}


\subsection{Text-conditioned Human Motion Synthesis}
%
Generative models have shown promising results on human motion synthesis~\cite{tevet2022human, chen2023executing, guo2024momask, dai2024motionlcm, zhang2025motion, zhao2024dartcontrol, wan2024tlcontrol}, though most works focus on single-agent or dual-agent scenarios.
%
%
%
Most recently, MaskControl~\cite{Pinyoanuntapong2025MaskControl} introduces accurate single-person controllability to the generative masked motion model~\cite{guo2024momask}, while maintaining high-quality generation.
%
%
%
Dual-agent motion synthesis has also seen rapid advancements~\cite{liang2024intergen, ponce2024in2in, xu2024inter}.
%
%
\citet{ma2025intersyn} employs an interleaved learning strategy to capture the dynamic interactions and nuanced coordination, exhibiting higher text-to-motion alignment, and improved diversity.
\citet{wang2025timotion} subsequently introduces TIMotion, a parameter-efficient approach utilizing temporal modeling and interaction mixing.
%
Synthesizing human-like reactions~\cite{tan2025think} is another active area of research.
%
\citet{xu2024regennet} establishes one of the earliest multi-setting benchmarks for this task, supported by three dedicated annotated datasets.
%
Similar to us, \citet{jiang2025arflow} propose direct noise-free action-to-reaction mappings through flow matching, while they ignore the error accumulation for autoregressive multi-person generation.
%

\vspace{-2mm}
\subsection{Unified Motion Synthesis}
The recent success of Large Language Models~\cite{achiam2023gpt, guo2025deepseek, agarwal2025cosmos, chen2024motionclr}, particularly their strong generative and zero-shot transfer capabilities, has inspired new generalist approaches in motion synthesis.
Research in unified motion generation has focused on several aspects, including: 1) unifying generation with understanding~\cite{zhu2025motiongpt3, jiang2023motiongpt}, 2) integrating diverse input modalities~\cite{li2025genmo, petrov2024tridi}, and 3) handling a variable number of actors~\cite{gupta2025unified, fan2025freemotion, zhao2025freedance}.
%
%
An early work~\cite{jiang2023motiongpt} proposed MotionGPT to address diverse motion-relevant tasks, which treats human motion as a foreign language to unify tasks like motion generation and understanding.
Then, \citet{petrov2024tridi} proposed TriDi for human-object interaction, a unified model capturing the joint 3D distribution of humans, objects, and their interactions.
%
To unify motion generation across different conditioning modalities (\eg, text, video), \citet{li2025genmo} introduced GENMO, a generalist model conditioned on videos, music, text, 2D keypoints, and 3D keyframes.
~\cite{gupta2025unified} introduced dualFlow, a flow-based model for interactive and reactive text-to-motion, though it is limited to dual-agent scenarios.
Most related to our work, FreeMotion~\cite{fan2025freemotion} proposes a decoupled generation and interaction module for number-free motion generation, while it suffers from inefficiency and error accumulation in multi-person scenarios.
%
Recently, \citet{zhao2025freedance} proposed FreeDance, a unified, number-free music-to-motion framework based on masked modeling of 2D discrete tokens, whereas our UMF focuses on the text-to-motion task.

\begin{figure*}[ht] 
\centering    
\includegraphics[width=0.85\textwidth]{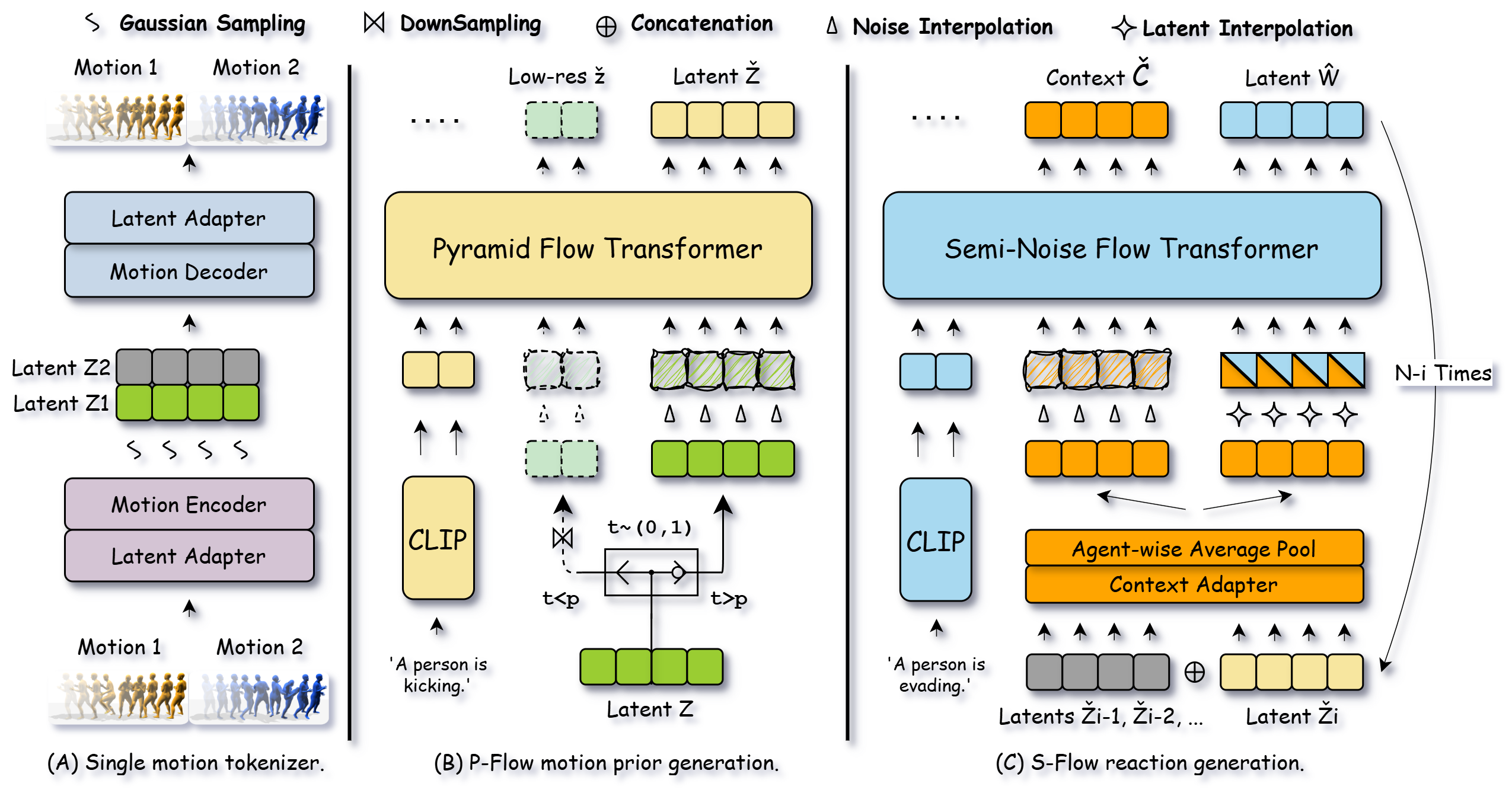}
\vspace{-5mm}
\caption{
\textbf{Overview of the Unified Motion Flow (UMF) architecture.}
The UMF framework consists of three stages.
\textbf{(A) Unified motion VAE:} A motion VAE with latent adapters encodes raw motions from heterogeneous datasets (e.g., HumanML3D~\cite{guo2022generating}, InterHuman~\cite{liang2024intergen}) into a regularized multi-token latent representation ($Z$).
\textbf{(B) P-Flow motion prior generation:} The Pyramid Flow Transformer synthesizes the latent motion prior ($\check{Z}$) based on noisy latent motion and text conditions. 
The P-Flow operates hierarchically based on the timestep $t \sim (0, 1)$: it processes downsampled, low-resolution latents for $t<p$ and switches to original-resolution latents for $t>p$, mitigating multi-token computational overheads.
\textbf{(C) S-Flow reaction generation:} 
Based on the previously generated latent \{$\check{Z}_i, \dots, \check{Z}_1$\}, the context adapter generates the context motion $C$.
Then the Semi-Noise Flow transformer predicts the reaction latent ($\check{W}$) by jointly modeling context reconstruction and reaction transformation, alleviating the error accumulation from previously generated motion.
}
\label{main_archi}
\vspace{-5mm}
\end{figure*}

\section{Preliminaries}
\label{fm_pre}

\textbf{Flow Matching.}
Flow generative models \cite{lipman2022flow, liu2022flow, albergo2022building} aim to learn a velocity field $v_t$ that maps source distribution $x_0 \sim p$ to target distribution $x_1 \sim q$ via an ordinary differential equation (ODE):
\small
\begin{equation}
\frac{dx_t}{dt} = v_t(x_t).
\label{eq:ode}
\end{equation}
\normalsize
Recently, \citet{lipman2022flow} proposed the flow matching framework, which offers a simulation-free training objective by directly regressing the model's velocity field $v_t$ on a conditional vector field $u_t(\cdot|x_1)$:
\small
\begin{equation}
\mathbb{E}_{t,q(x_1),p_t(x_t|x_1)} \left\| v_t(x_t) - u_t(x_t|x_1) \right\|^2,
\label{eq:flow_matching_objective}
\end{equation}
\normalsize
where $u_t(\cdot|x_1)$ uniquely determines a conditional probability path $p_t(\cdot|x_1)$ toward data sample $x_1$. An effective choice of the conditional probability path is linear interpolation~\cite{ma2024sit} of data and noise:
\small
\begin{equation}
x_t = tx_1 + (1-t)x_0,
\label{eq:linear_interpolation}
\end{equation}
\begin{equation}
x_t \sim \mathcal{N}(tx_1, (1-t)^2 I),
\label{eq:interpolation_distribution}
\end{equation}
\normalsize
and $u(x_t|x_1) = x_1 - x_0$. Notably, flow matching can be flexibly extended to interpolate between distributions other than Gaussians. This enables us to employ the flow matching for both motion prior and reaction generation.

\section{Proposed Method}

\subsection{Unified Latent Space}
%

A key challenge in building a generalist motion model is that generative frameworks like flow matching require a consistent data format, a condition not met by heterogeneous motion datasets. For instance, individual motion datasets \cite{guo2022generating} often use canonical representations, while interaction datasets \cite{liang2024intergen} use non-canonical representations. To bridge this gap, we first convert individual motions to a unified non-canonical SMPL skeleton representation with 22 joints.
Then we split the interaction sample into multiple individual motion sequences (see Appendix A for details).

As shown in Fig.~\ref{main_archi}(A), the single motion tokenizer learns a continuous latent space for individual motion sequences.
Similar to TEMOS~\cite{petrovich2022temos}, we utilize transformers~\cite{vaswani2017attention} as the encoder and decoder, enhanced with skip connections and layer norms.
The individual encoder takes an individual motion sequence $x^{1:N}_{I} \in \mathbb{R}^{N \times D}$ as input and compresses it into a latent representation $z \in \mathbb{R}^{p \times r}$.
Using the reparameterization trick~\cite{kingma2013auto}, we sample a latent vector $z \in \mathbb{R}^{p \times r}$ from the learned Gaussian distribution. Then, the individual decoder reconstructs the latent vector $z$ into motion sequences $\hat{x}^{1:N}_{I}$.
%
Different from existing number-free methods~\cite{fan2025freemotion} that are trained on raw motion space, which suffer from performance degradation on heterogeneous datasets, our multi-token latent space shows better stability.
%
%

\textbf{Multiple latent tokens.}
Previous latent motion diffusion works~\cite{chen2023executing, zhu2025motiongpt3} employ \textit{single latent token} learning (\eg $1\times 256$), imposing a bottleneck on the VAE's reconstruction performance.
%
While naively increasing the number of tokens can improve reconstruction, it often degrades the generative performance~\cite{yao2025reconstruction}.
Inspired by~\citet{motionlcm-v2}, we utilize a latent adapter
to decouple the internal token representation from the final latent dimension.
The VAE encoder first captures complex motion details using a larger token (e.g., $16 \times 256$) and then projects them to a compact, semantically dense space (e.g., $16 \times 32$) for the motion generation.
This design achieves a better trade-off between reconstruction capacity and generative quality (See Sec.~\ref{abs1}).

\textbf{Regularized latent space.}
In a typical VAE training process, motion reconstruction $x^{1:N}$ is constrained by the Mean Squared Error (MSE) and Kullback-Leibler (KL) losses. 
%
%
We further adapt the geometric loss~\cite{tevet2022human}, which enhances the physical plausibility within involved individuals and preserves the original interaction relationships between individuals. 
The training loss of VAE is:
\vspace{-2mm}
\begin{equation}
\mathcal{L}_\text{VAE}= 
\scalebox{1}{$
\mathcal{L}_{\text{geometric}}  + \mathcal{L}_{\text {reconstruction }} + \lambda_{\text {KL }} \mathcal{L}_{\text {KL }}. 
$}
\end{equation}

%

\subsection{Unified Motion Flow Matching}
%

As shown in Fig.~\ref{main_archi}, based on the multi-token latent space, we decouple the number-free motion generation process into two stages:
(1) Motion Prior Generation: An individual motion prior is generated via the \textbf{Pyramid Motion Flow (P-Flow)}, a hierarchical flow matching process conditioned on the timestep.
Unlike Denoising Diffusion Probabilistic Models (DDPMs)~\cite{ho2020denoising} operating in the raw motion space, this design offers better scalability \cite{esser2024scaling} and efficiency within multi-token latent spaces~\cite{jin2024pyramidal, ran2025tpdiff}.
(2) Reaction Motion Generation: Given the motion prior (or preceding reaction), \textbf{Semi-Noise Motion Flow (S-Flow)} learns a joint path for context reconstruction and reaction transformation for the next person.
Instead of fine-tuning complex ControlNet \cite{zhang2023adding}, S-Flow learns an adaptive, context-aware motion transition, alleviating potential error accumulation.

%

\textbf{Scalability to Group Scenarios ($N>2$).}
Due to the scarcity of SMPL-based~\cite{SMPL:2015} datasets featuring $\geq$3 interacting agents, our framework is mainly trained and evaluated on dual-agent scenarios, while UMF is not limited to this setting.
For $N>2$ people, the S-Flow module is applied autoregressively, using the synthesized motions of preceding agents as input to generate the next agent's motion.
We demonstrate its zero-shot capability via a user study (Sec.~\ref{qr_us}).

\subsubsection{Motion Prior}
Compared to single-token approaches, the multi-token latent space unlocks better motion generation conditioned on text prompt $c$, but it also imposes more computational demands.
A key observation is that initial generation steps~\cite{wang2025lavie} often operate on noisy and less informative variables, suggesting that the entire full resolution is not necessary.
Previous works address this by training multiple models with different resolutions~\cite{xie2024towards, jiang2025motionpcm} based on the timestep, which still introduces extra model complexity.
We introduce the Pyramid Motion Flow (P-Flow) \cite{jin2024pyramidal}, which reinterprets the Gaussian flow matching trajectory as hierarchical stages within one transformer model.
Each stage operates at a resolution corresponding to the timestep, where only the final stage uses the original resolution, enabling efficient flow matching inference.

\textbf{P-Flow forward process.}
%
Unlike standard Gaussian flow matching~\cite{lipman2022flow, hu2023motion} that evolves between full-resolution noise and data, P-Flow starts with a coarser interpolation between downsampled latent motion, and progressively yields finer-grained, higher-resolution endpoints.
%
To handle the varying dimensions of $z_t$, we decompose the trajectory into a piecewise flow~\cite{yan2024perflow}. It divides $[0, 1]$ into $K$ time windows, each interpolating between successive resolutions with a unique start and end point.
For the $k$-th time window $[s_k, e_k]$, we jointly compute the endpoints $(\hat{z}_{s_k}, \hat{z}_{e_k})$ with noise $\epsilon \sim \mathcal{N}(\mathbf{0}, I)$ and data point $z_1$ as:
\small
\begin{align}
\text{Start Point:} \quad &\hat{z}_{s_k} = s_k Up(Down(z_1, 2^{k})) + (1 - s_k)\epsilon, \label{eq:coupled_start} \\ 
\text{End Point:} \quad &\hat{z}_{e_k} = e_k Down(z_1, 2^{k-1}) + (1 - e_k)\epsilon, \label{eq:coupled_end}
\end{align}
\normalsize
where $k\in [K, 1]$, $Up(\cdot)$ and $Down(\cdot)$ are standard resampling functions and irreversible between them.
Notably, $Up(Down(z, 2^1))$ is a lossy approximation of $z$, which forces the flow model to learn the correlation between resolutions.
The path spans from pure noise $\epsilon$ (at $k=K, s_k=0, \hat{z}_{s_k} = \epsilon$) to the data point $z_1$ (at $k = 1, e_k=1, \hat{z}_{e_k} = Down(z_1, 2^0) = z_1$).
%

%
To enhance the straightness of the flow trajectory, we couple the sampling of its endpoints by enforcing the noise $\epsilon$ to be in the same direction. 
%
%
Let $t' = (t - s_k)/(e_k - s_k)$ denote the rescaled timestep, then the flow within it follows:
\small
\begin{equation}
\hat{z}_t = t'\hat{z}_{e_k}+ (1 - t') \hat{z}_{s_k},
\label{eq:piecewise_flow}
\end{equation}
\normalsize
Here, the trajectory at $k$-th stage starts at $\hat{z}_{s_k}$ and ends at $\hat{z}_{e_k}$.
%
%
This pyramidal structure, applicable to spatial or temporal dimensions, concentrates computation at lower resolutions, reducing the cost by a factor of $\approx 1/K$ in theory.
%

Thereafter, we can regress the flow model $G^P_\theta$ on the conditional vector field $u_t(\hat{z}_t|z_1) = \hat{z}_{e_k} - \hat{z}_{s_k}$ with the following objective to unify different stages:
\small
\begin{equation}
\mathcal{L}_\text{P-Flow}= \mathbb{E}_{k,t,\hat{z}_{e_k},\hat{z}_{s_k}} \left\| G^P_\theta(\hat{z}_t; t, c) - (\hat{z}_{e_k} - \hat{z}_{s_k}) \right\|^2.
\label{eq:unified_flow_matching}
\end{equation}
\normalsize

\textbf{P-Flow sampling process.} 
%
Using Euler ODE solvers, each pyramid stage is discretized into $M=T_{P_k}$ steps:
\small
\begin{equation}
\vspace{-1mm}
\hat{z}_{t_{m+1}} \leftarrow \hat{z}_{t_m} + (t_{m+1} - t_m) G^P_\theta(\hat{z}_{t_m}, t_m, c),
\label{odep}
\vspace{-1mm}
\end{equation}
\normalsize
where $t_1 = s_k, \cdots, t_M = e_k$ are the discrete timesteps.
However, we must carefully handle the jump points~\cite{campbell2023trans} between successive pyramid stages of different resolutions to ensure continuity of the probability path.
%

As shown in Algorithm 1, for the transition from stage $k$ to $k-1$, we first upsample the previous endpoint $\hat{z}_{e_{k}}$ via nearest-neighbor interpolation.
The inference has to match the Gaussian distributions at each jump point by a linear transformation of the upsampled result. 
Specifically, the following rescaling and renoising scheme suffices:
\small
\begin{equation}
\vspace{-1mm}
\hat{z}_{s_{k-1}} = \frac{s_{k-1}}{e_{k}} Up(\hat{z}_{e_{k}}) + \alpha n', \quad \text{s.t. } n' \sim \mathcal{N}(0, \Sigma'),
\label{eq:rescaling_renoising}
\vspace{-1mm}
\end{equation}
\normalsize
where $\Sigma'$ is a blockwise diagonal covariance matrix (e.g., $4 \times 4$ blocks).
The coefficient $s_{k-1}/e_{k}$ matches the means, and the corrective noise $\alpha n'$ matches the covariances.
To ensure continuity after upsampling (see Appendix B for derivation), we set $e_{k} = 2s_{k-1}/(1 + s_{k-1})$ and $\alpha = \frac{\sqrt{3}(1 - s_{k-1})}{2}$ for a consistent mean and covariance.
%
%



%
%

\subsubsection{Reaction Motion Generation} 
%
%
For number-free motion generation, we generate the reaction $W$ conditioned on an arbitrary action (\ie, ${\hat{Z}_i}$) and text prompt $c$.
This process is applied iteratively to synthesize interactions involving more than two agents.
Based on the set $\mathcal{Z}_{gen}$ of previously generated motions,
Semi-Noise Flow (S-Flow) learns a \textbf{joint transformation} to generate reaction motion $W$ for subsequent characters, which is trained exclusively on the multi-person dataset.

As shown in Fig.~\ref{main_archi} (C), S-Flow reformulates reaction generation with context $C_i$ by adaptively optimizing two probability paths simultaneously: (1) \textbf{reaction transformation} (the path from $C_{i}$ to $W$) via context interpolation, and (2) \textbf{context reconstruction} (the path from $\epsilon$ to $C_i$) via Gaussian noise interpolation.
Instead of relying on complex conditional mechanisms like ControlNet \cite{xie2023omnicontrol, xu2024regennet, fan2025freemotion}, we first employ a context adapter to generate context motion, which is used as \textit{the direct input into flow matching}.
%
This design provides a more flexible starting point for learning the reaction transformation paths, allowing the adaptive adjustment for possibly sub-optimal motion from other characters.
The auxiliary context reconstruction path also helps S-Flow understand context at a global level, balancing its context-awareness and reaction forecasting, thereby alleviating overall error accumulation in autoregressive models.

%
%
%
%
%
%
\small
\begin{algorithm}[t]
\label{algo}
\caption{UMF Inference Algorithm}
\begin{algorithmic}[1]
\State \textbf{Input:} $c$ (text prompt); $N$ (agent number); $K$ (P-Flow stage count); Models ($G^P_\theta$, $G^S_\theta$, $Dec$, $TranEnc$)
\State \textbf{Parameters:} $\{T_{P_{k}}\}$ (P-Flow steps); $T_S$ (S-Flow steps)
\State \textcolor{gray}{\slshape \# P-Flow Motion Prior Generation}
\State $\hat{z}_{s_K} \sim \mathcal{N}(0, I), 0 \le s_k < e_k \le 1, s_{K} = 0, e_1 = 1$  
\For{$k = K$ \textbf{down to} $1$}
    \State $\hat{z}_{e_{k}} \leftarrow \text{SolveODE}(G^P_\theta, \hat{z}_{s_k}, c; T_{P_{k}})$ \Comment{Eq.~\ref{odep}}
    \If{$k \geq 2$}
        \State $\hat{z}_{s_{k-1}} \leftarrow \text{JumpUpdate}(\hat{z}_{e_{k}}, s_{k-1}, e_{k})$ \Comment{Eq.~\ref{eq:rescaling_renoising}}
    \EndIf
\EndFor
\State $\hat{Z}_1 \leftarrow \hat{z}_{e_1}$, $\mathcal{Z}_{gen} \leftarrow \{\hat{Z}_1\}$

\State \textcolor{gray}{\slshape \# S-Flow Reaction Generation}
\For{$i = 2$ to $N$}
    \State $C_i = TranEnc(\mathcal{Z}_{gen})$ \Comment{Context Adapter}
    \State $\hat{Z}_i \leftarrow \text{SolveODE}(G^S_\theta, C_i, c; T_S)$ \Comment{Eq.~\ref{odes}}
    \State $\mathcal{Z}_{gen} \leftarrow \mathcal{Z}_{gen} \cup \{\hat{Z}_i\}$  
\EndFor
\State \textcolor{gray}{\slshape \# VAE Decoding}
\State $\{x_1, \dots, x_N\} \leftarrow Dec(\mathcal{Z}_{gen})$
\State \textbf{Return} $\{x_1, \dots, x_N\}$
\end{algorithmic}
\end{algorithm}
\normalsize

\begin{figure*}[t]
    \centering
    
    {\setlength{\tabcolsep}{105pt}
    \begin{tabular}{*{8}{c}}
    \toprule
        \multicolumn{4}{c}{\large\bfseries UMF} & 
        \multicolumn{4}{c}{\large\bfseries FreeMotion} \\
        \bottomrule
    \end{tabular}
    }
    
    {\itshape \scriptsize{Two performers use their \textbf{right leg} to confront each other, and then one lifts the \textbf{left leg} to attack.}}
    {\setlength{\tabcolsep}{0pt}
    \begin{tabular}{*{8}{c}}
        \includegraphics[width=0.125\textwidth]{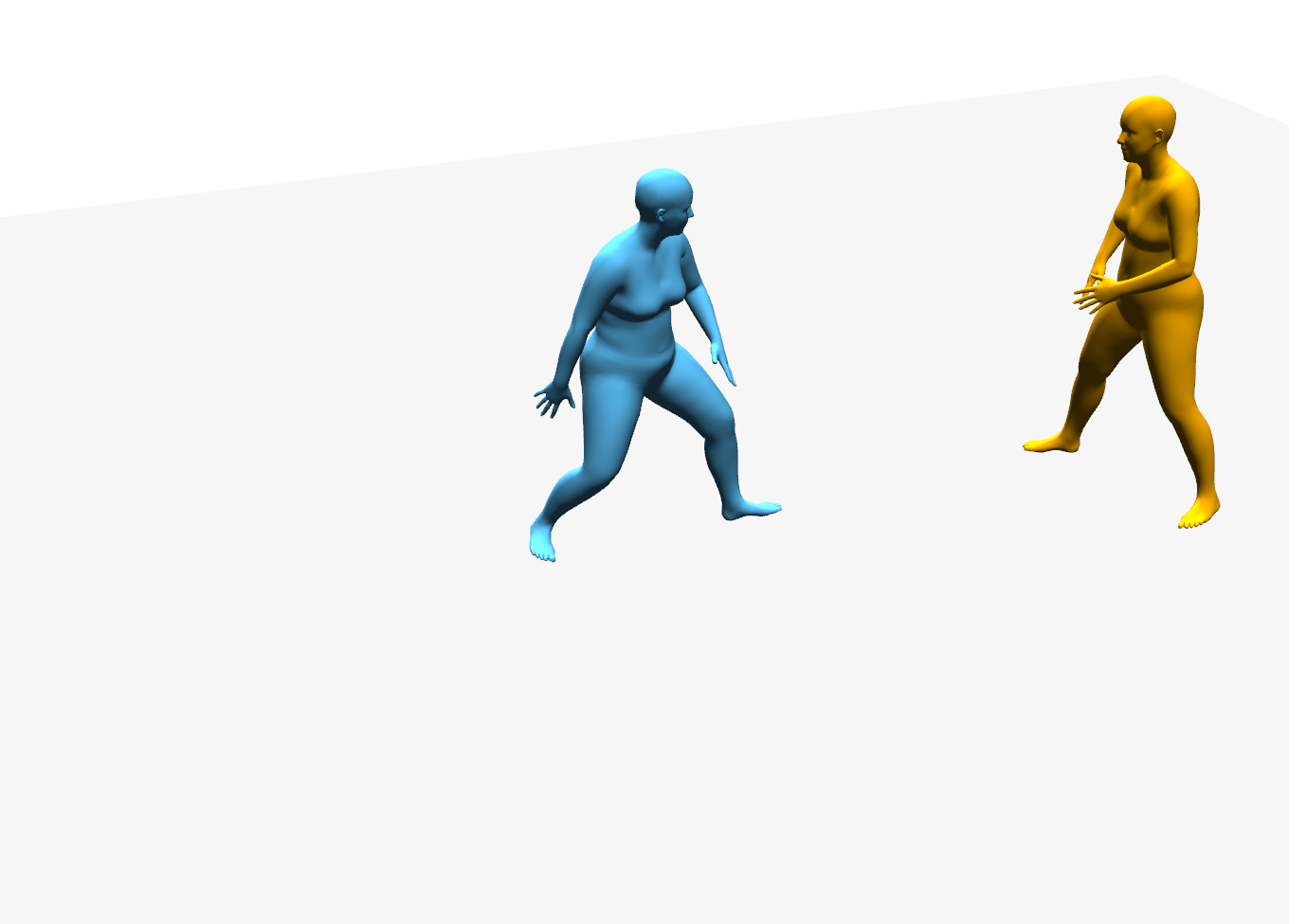} &
        \includegraphics[width=0.125\textwidth]{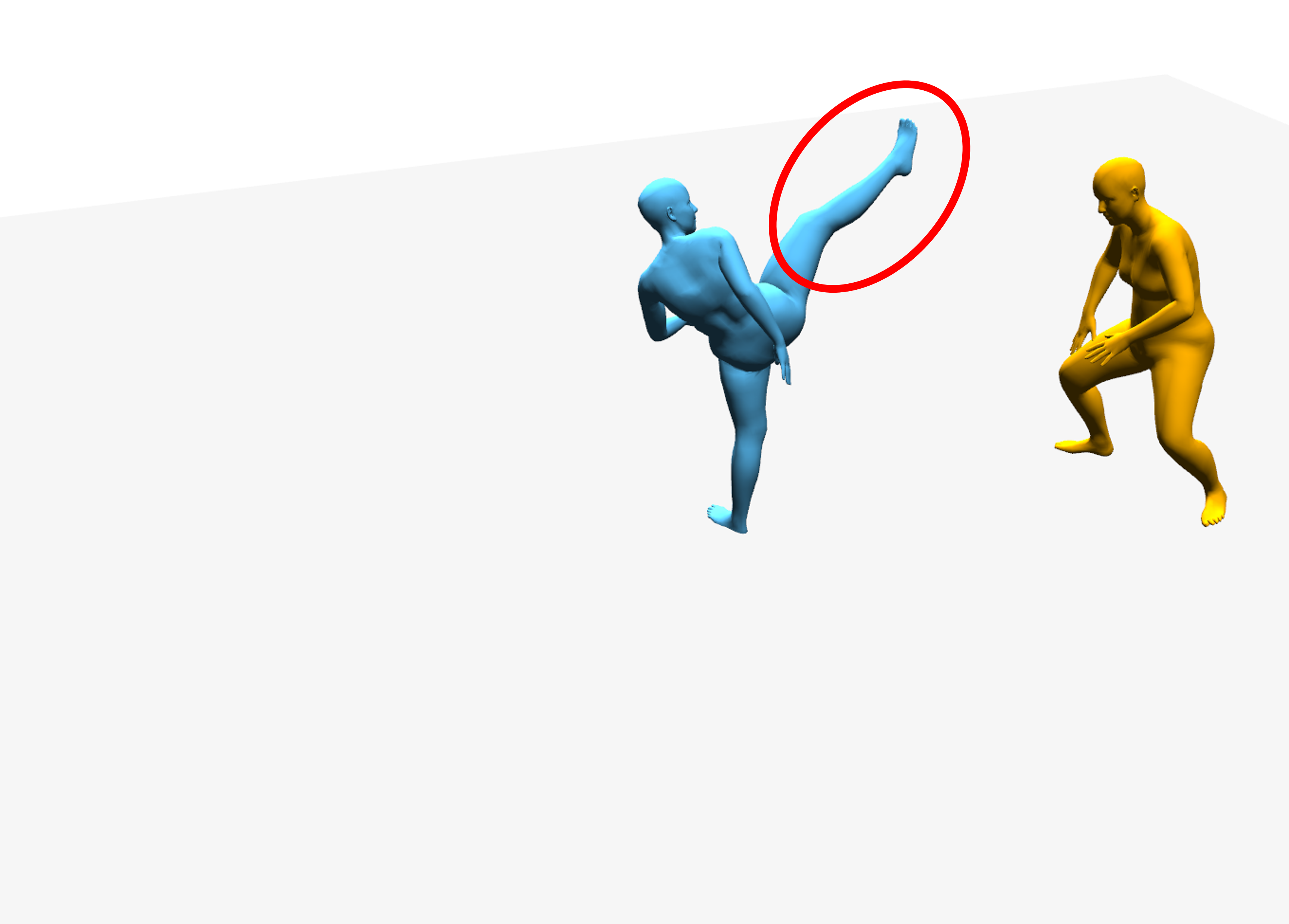} &
        \includegraphics[width=0.125\textwidth]{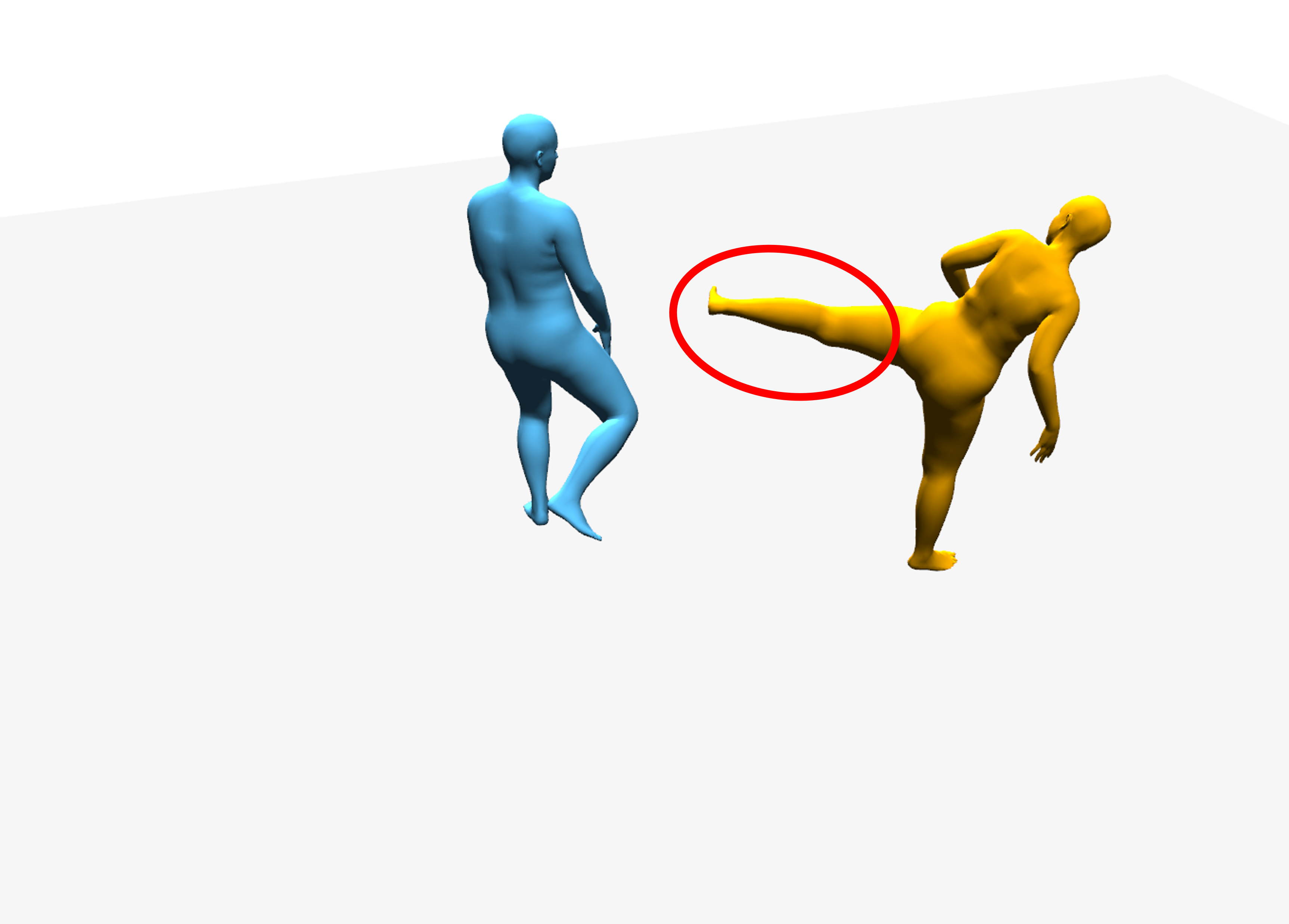} &
        \includegraphics[width=0.125\textwidth]{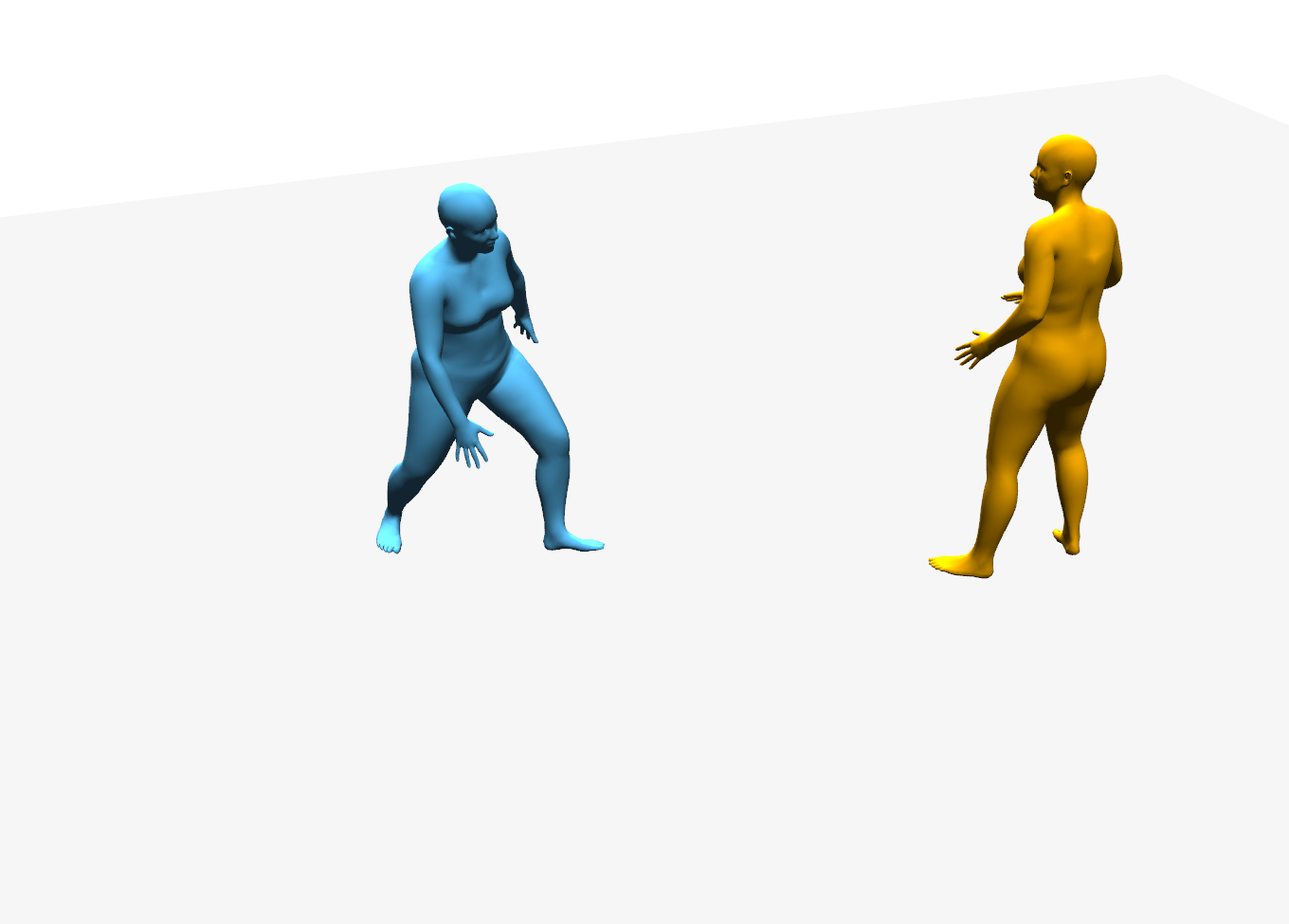} \hspace*{5pt}& 
        
        \includegraphics[width=0.125\textwidth]{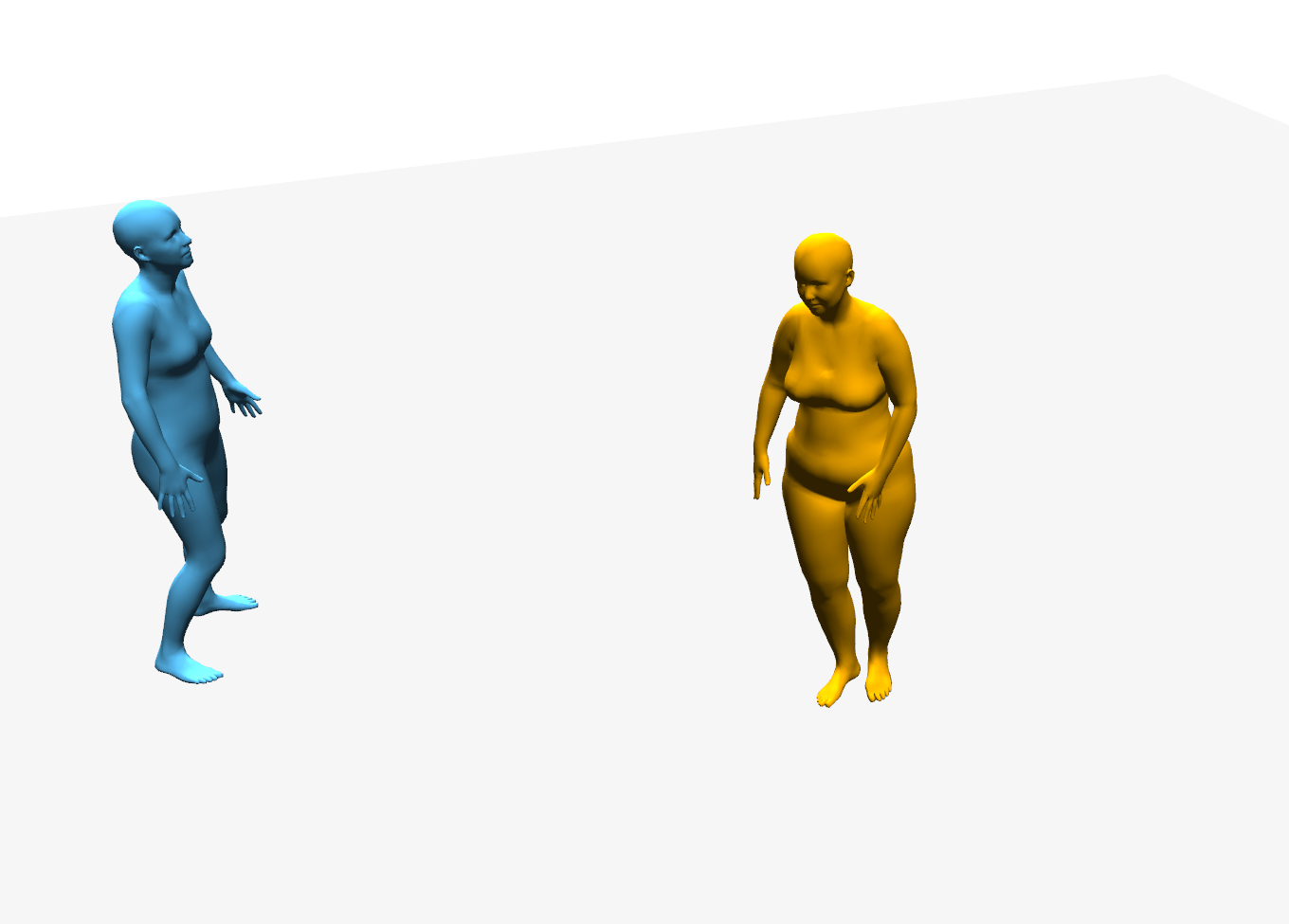} &
        \includegraphics[width=0.125\textwidth]{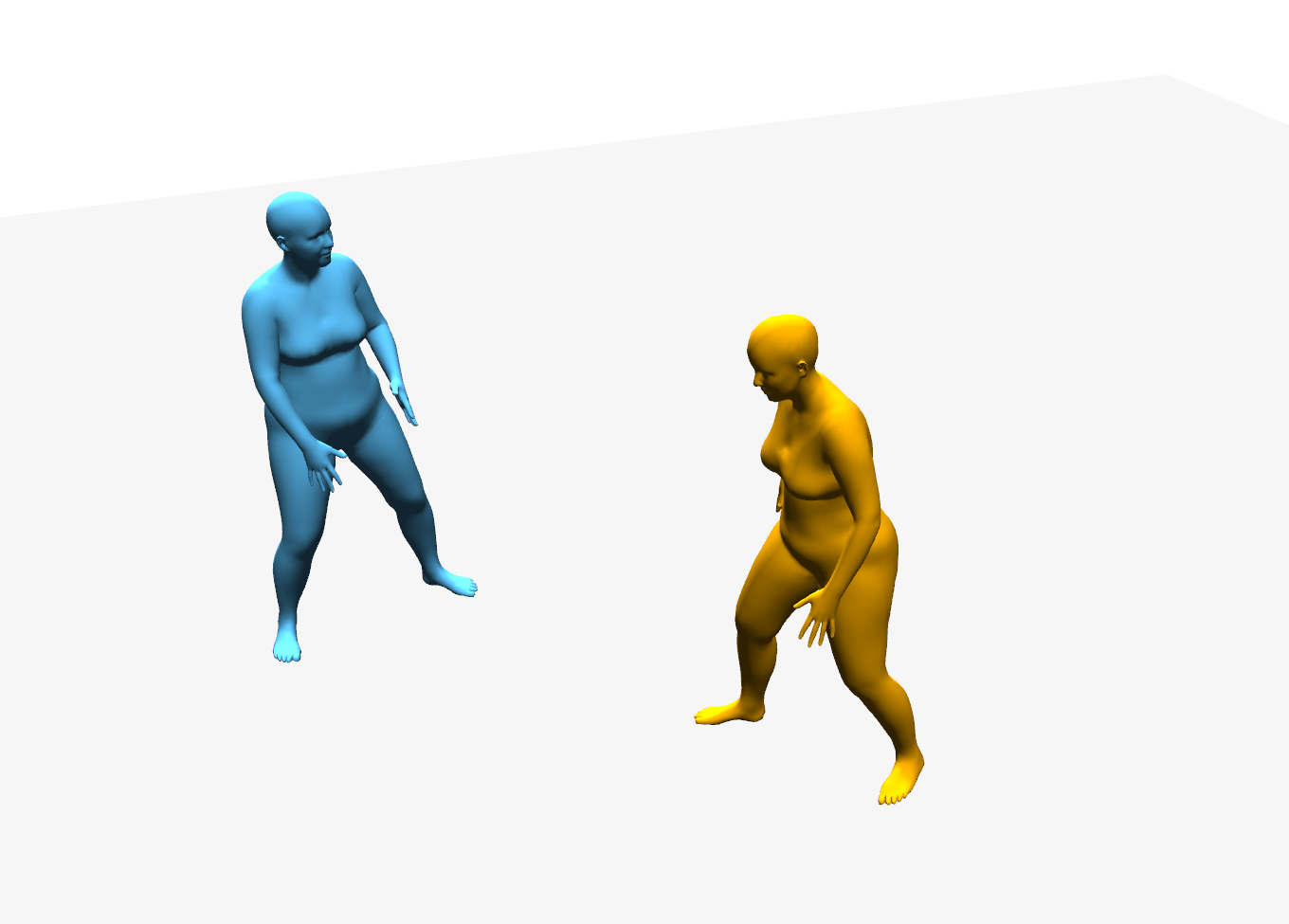} &
        \includegraphics[width=0.125\textwidth]{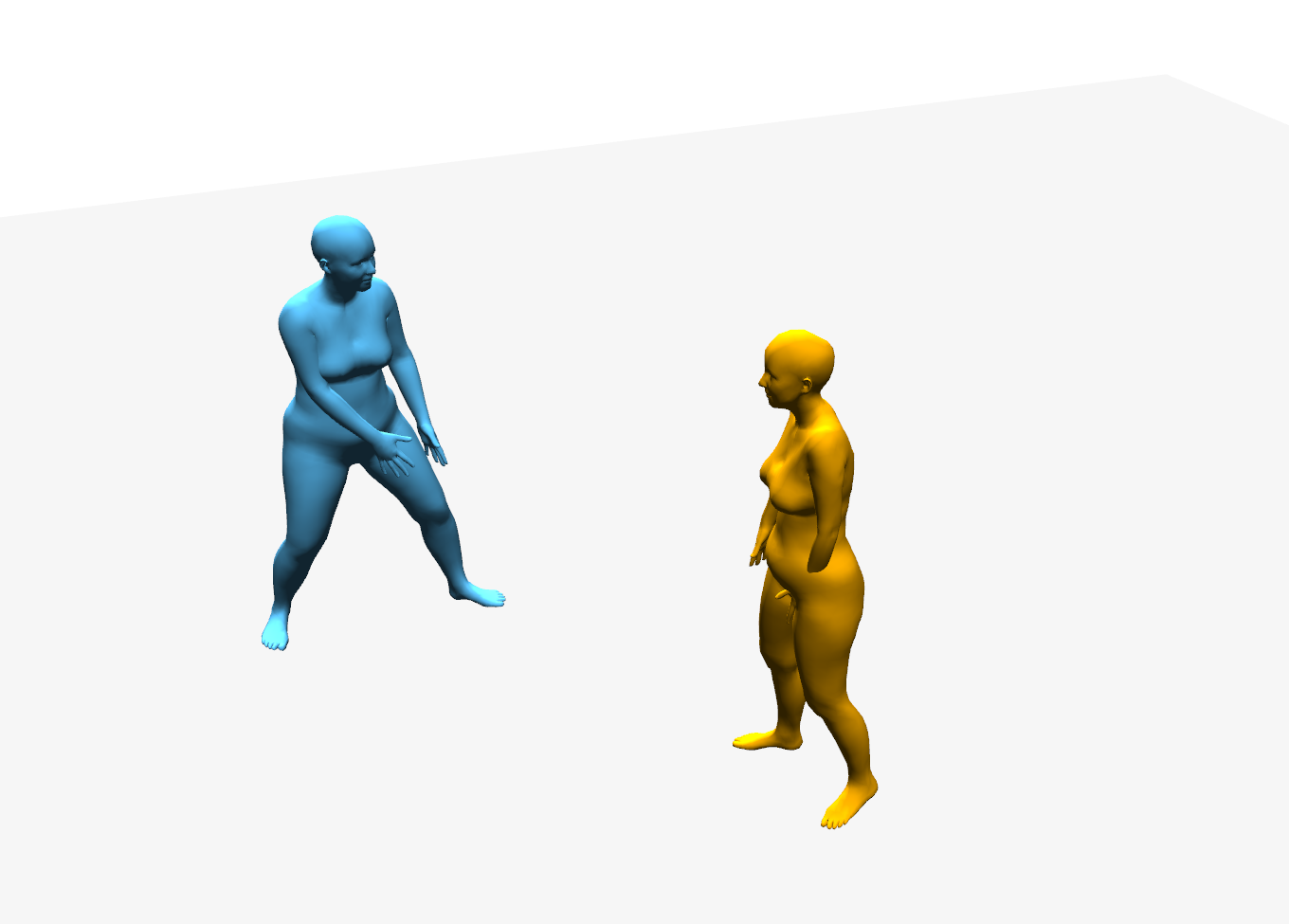} &
        \includegraphics[width=0.125\textwidth]{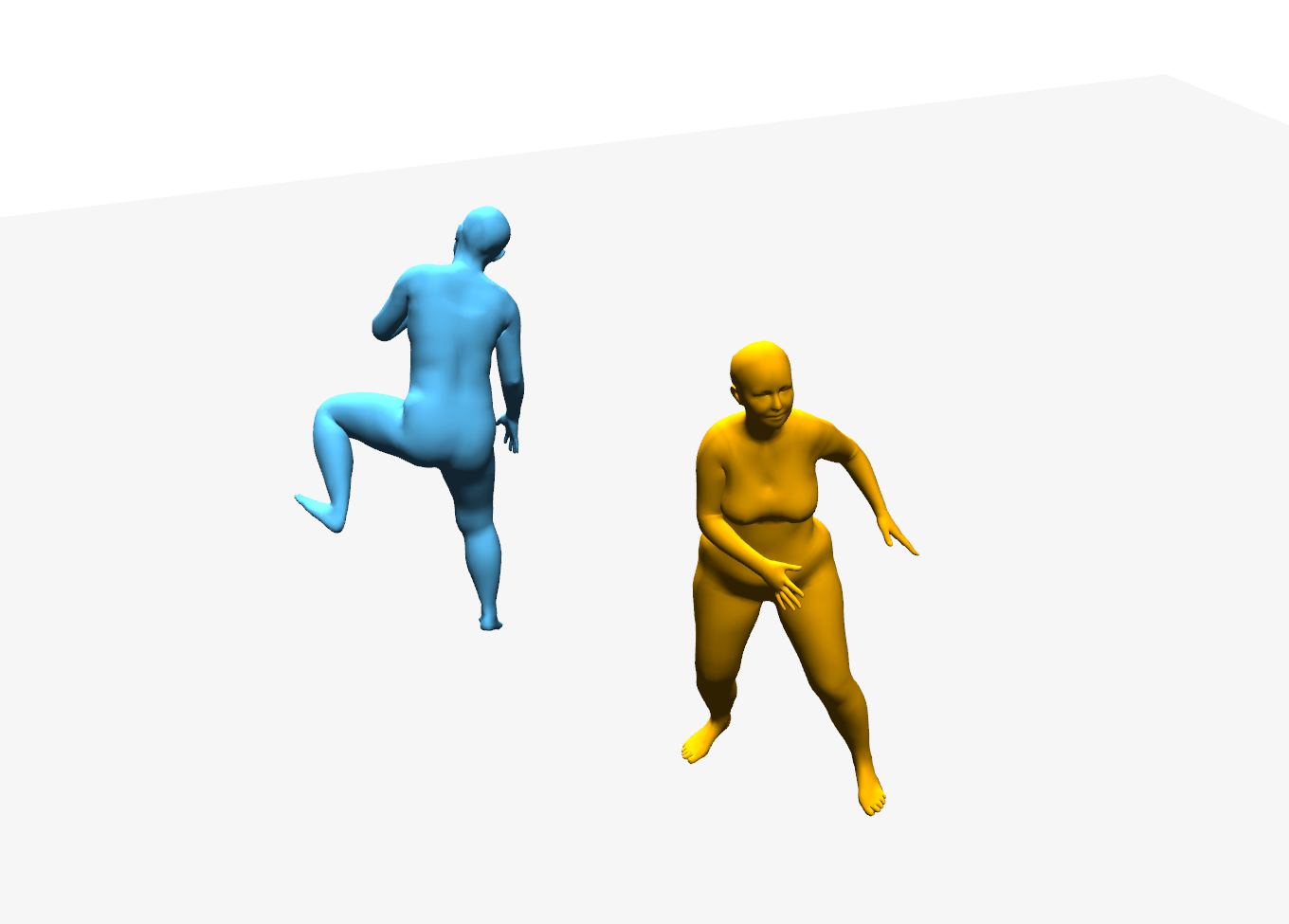} \\ 
    \end{tabular}}

        {\itshape \scriptsize{Two people stroll together and chatter with each other, the third person walks towards them with hand gestures, later they are \textbf{walking together}.}}
        {\setlength{\tabcolsep}{0pt}
    \begin{tabular}{*{8}{c}}
        \includegraphics[width=0.125\textwidth]{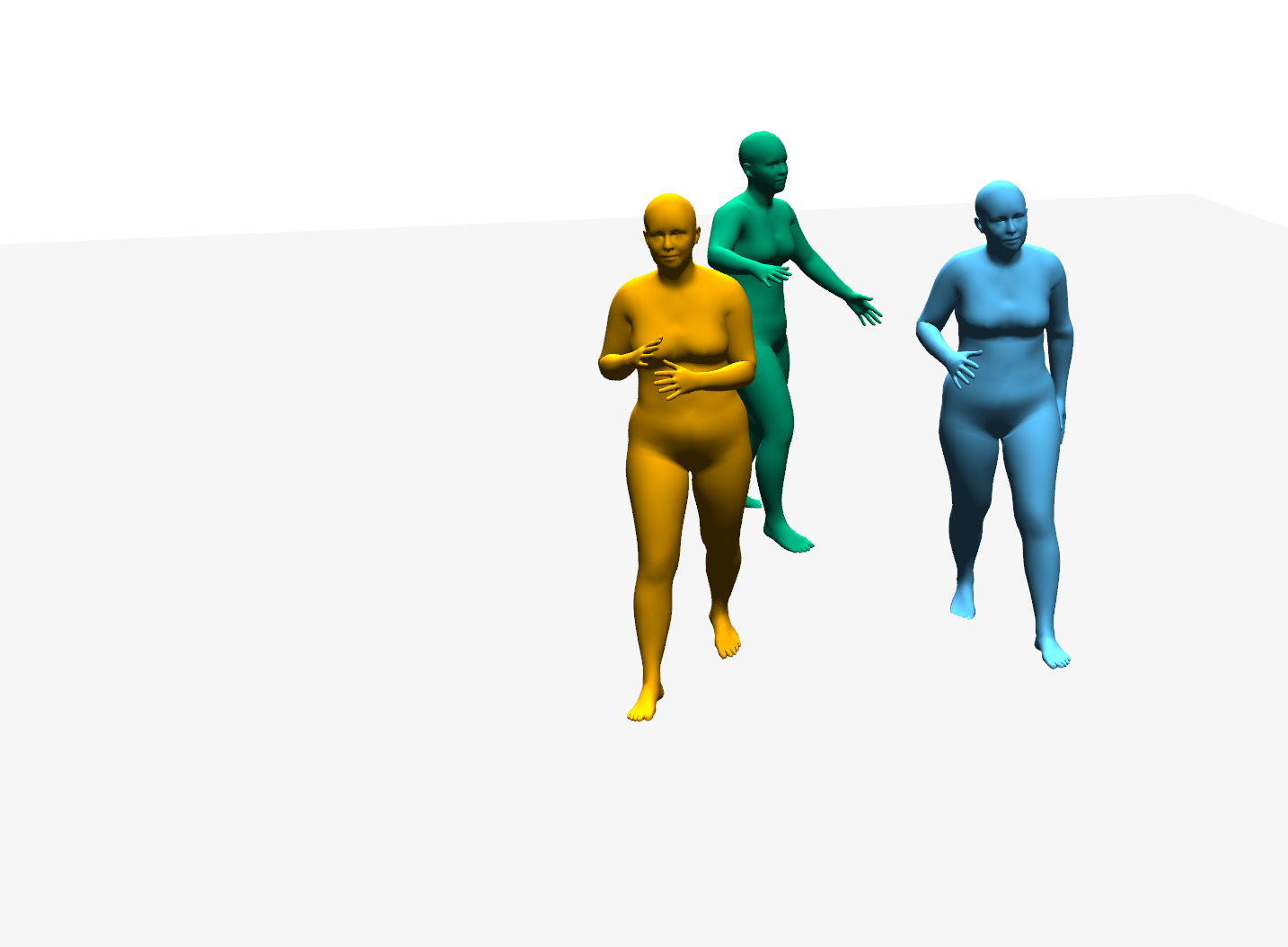} &
        \includegraphics[width=0.125\textwidth]{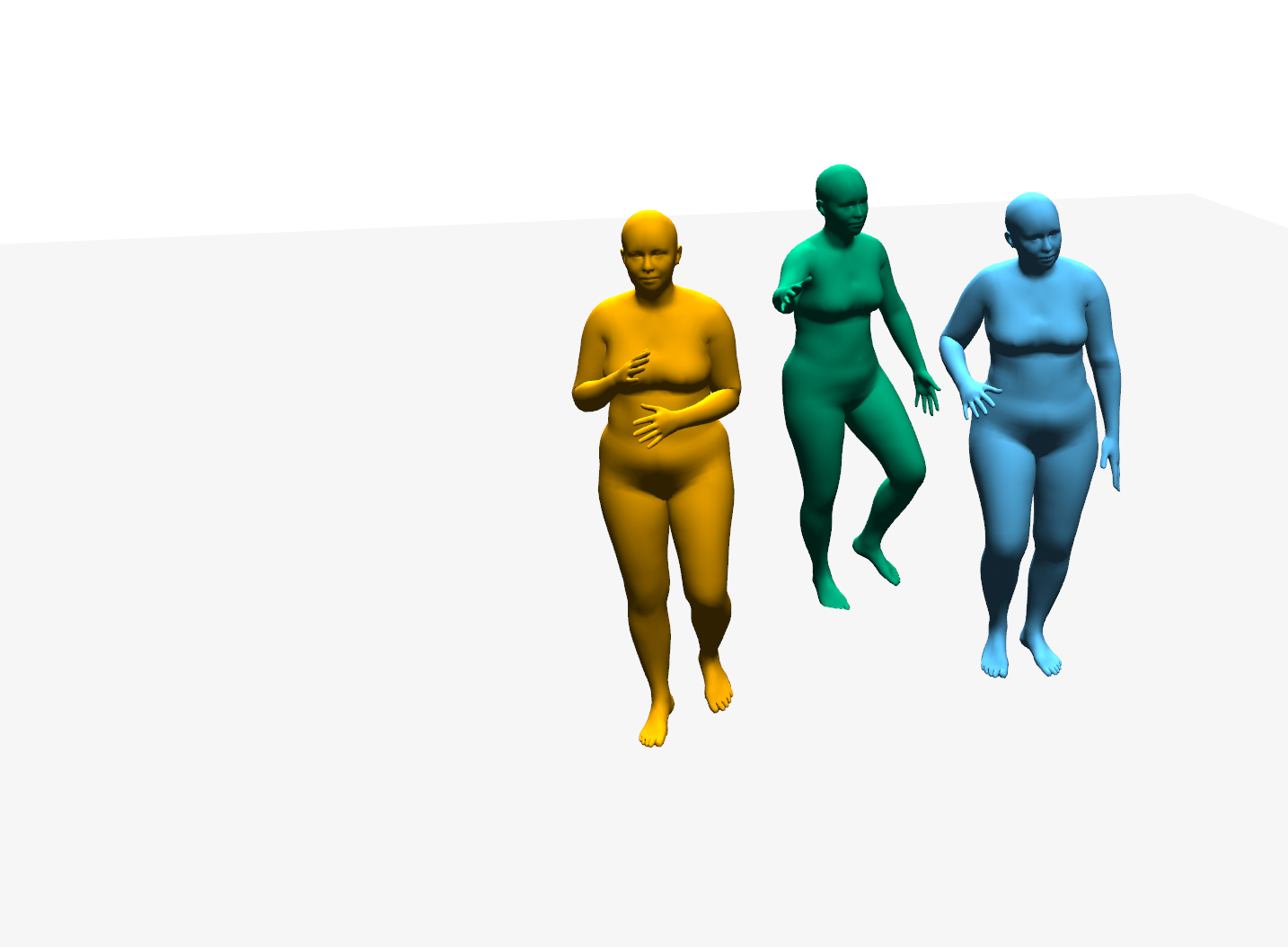} &
        \includegraphics[width=0.125\textwidth]{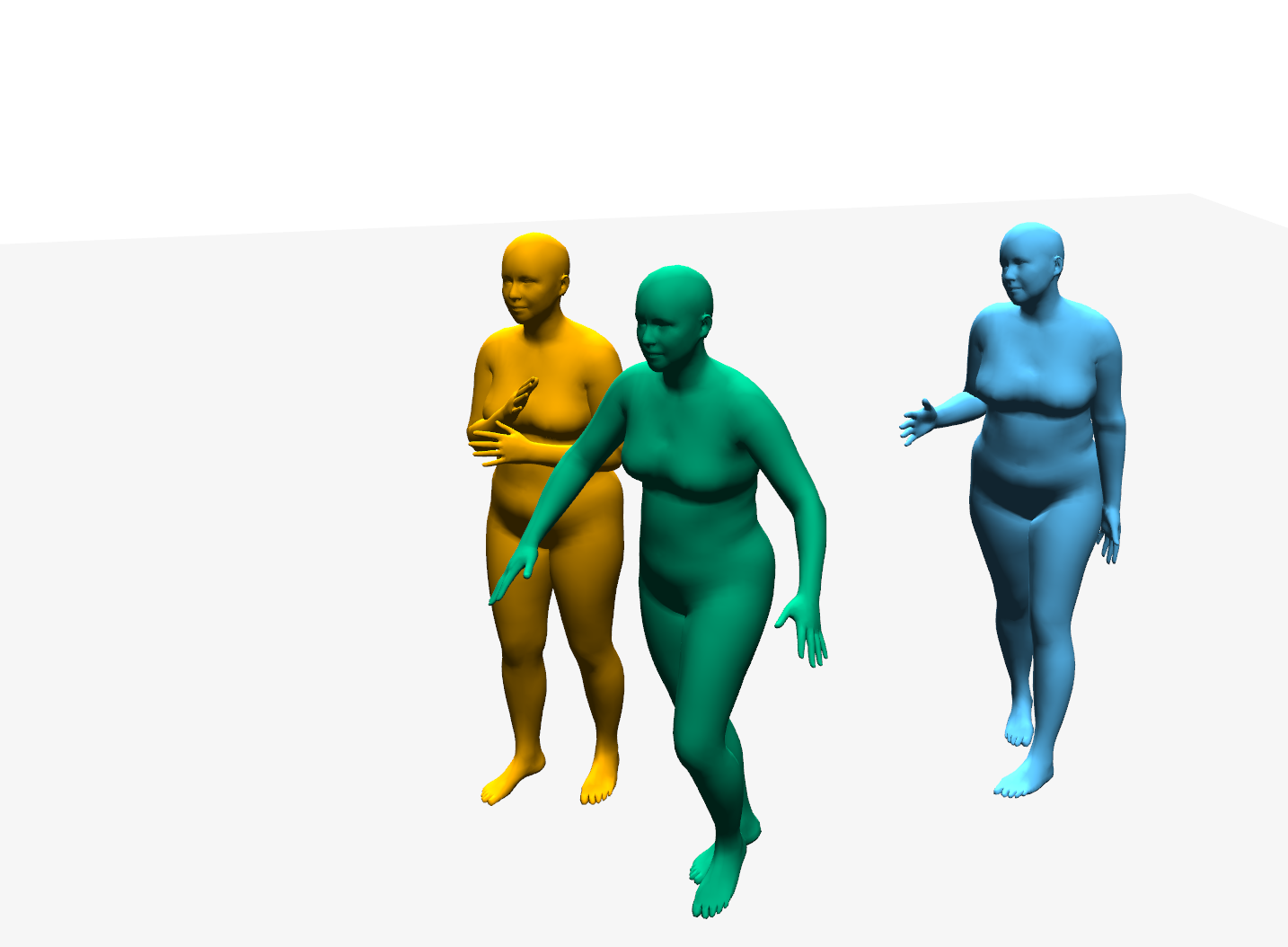} &
        \includegraphics[width=0.125\textwidth]{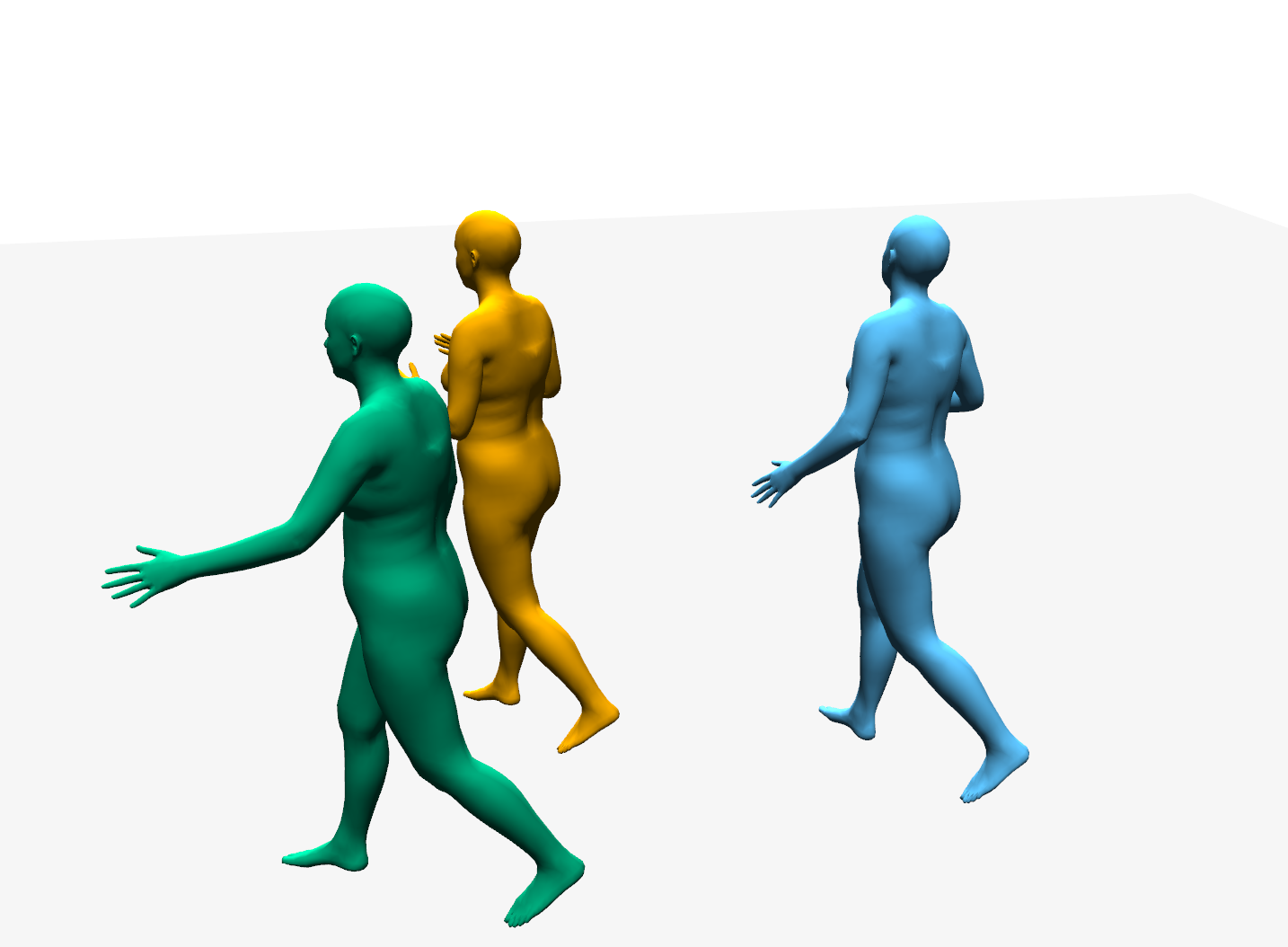} \hspace*{5pt}& 
        
        \includegraphics[width=0.125\textwidth]{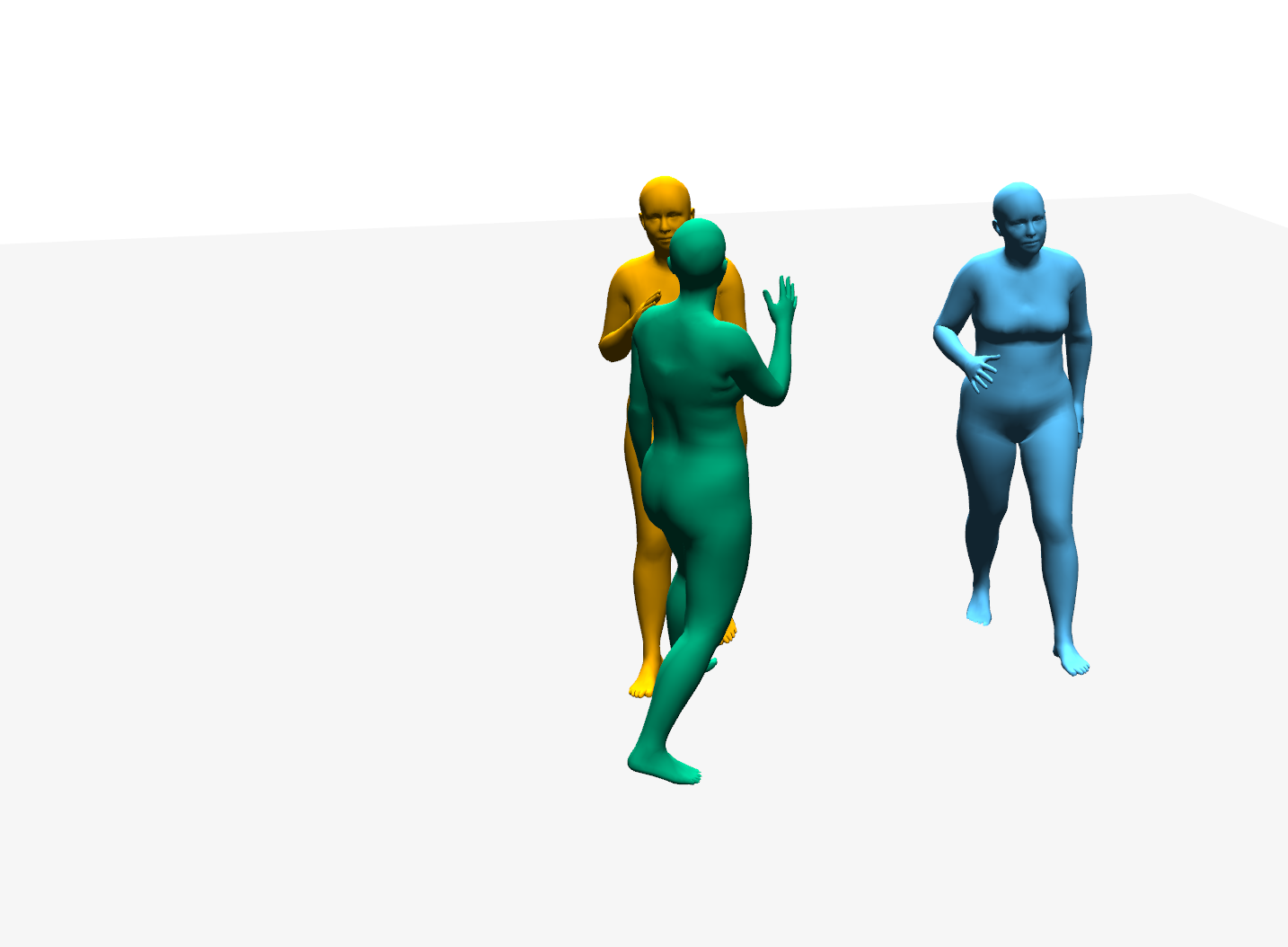} &
        \includegraphics[width=0.125\textwidth]{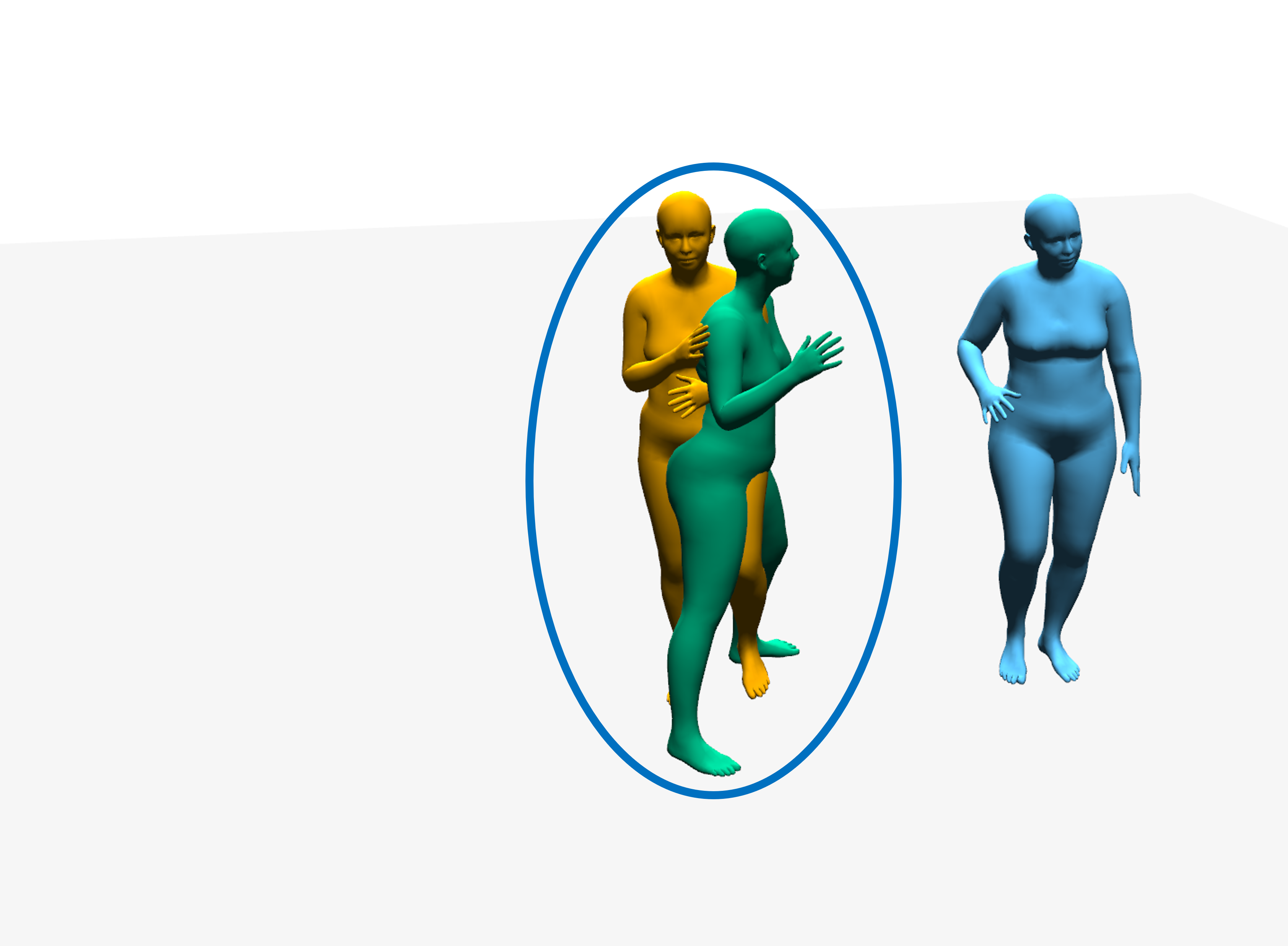} &
        \includegraphics[width=0.125\textwidth]{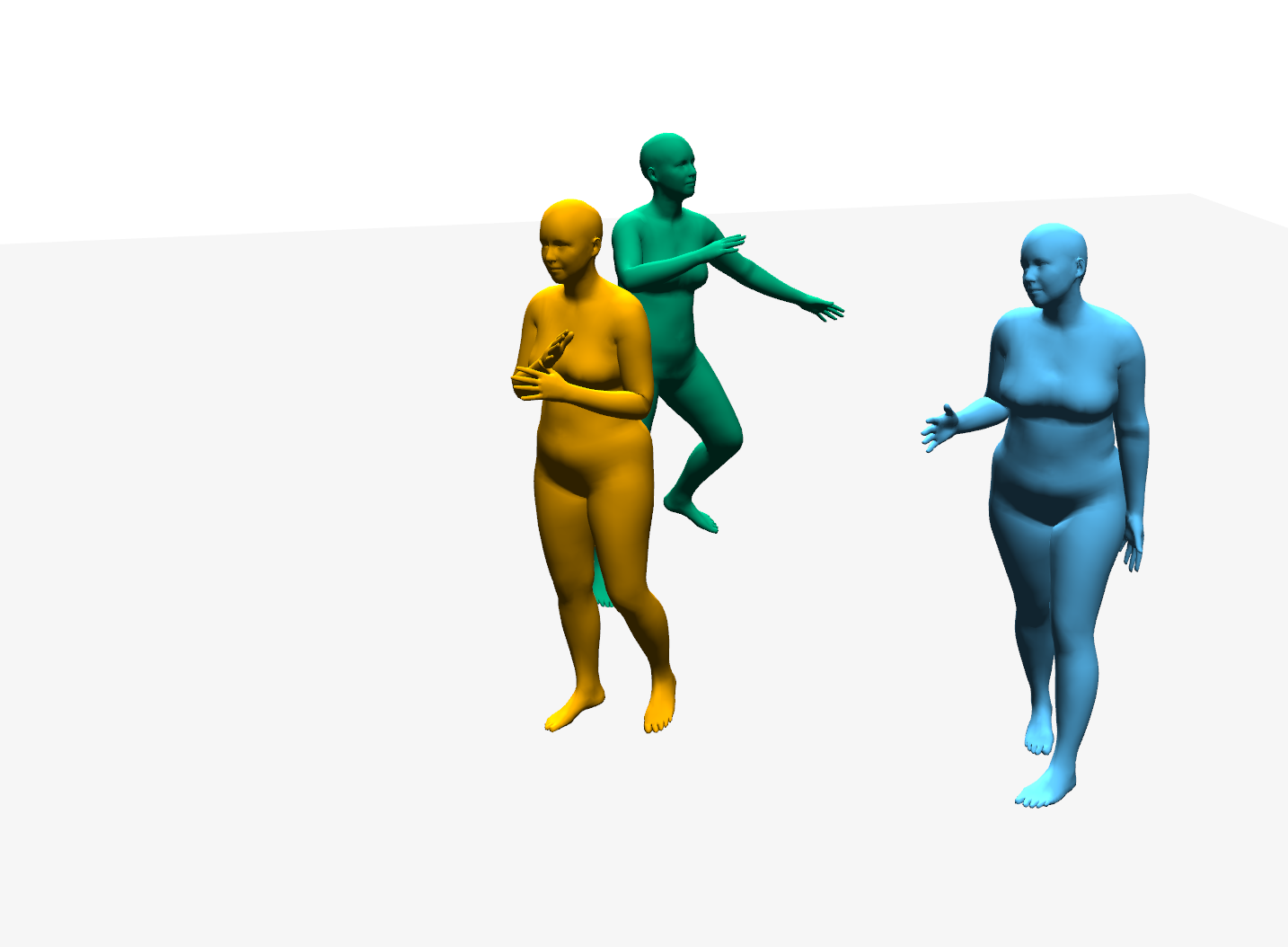} &
        \includegraphics[width=0.125\textwidth]{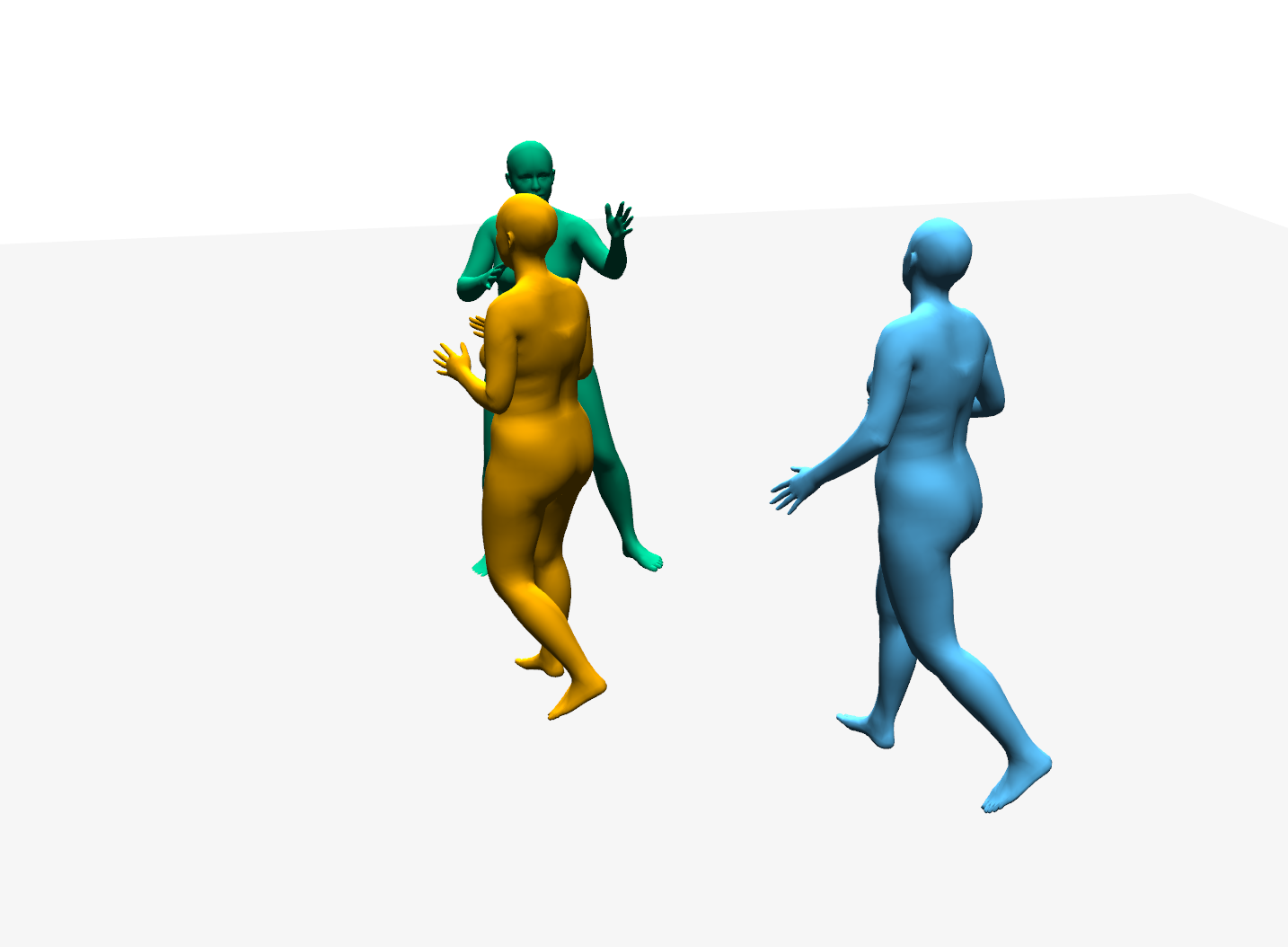} \\ 
    \end{tabular}}

        {\itshape \scriptsize{Two people are sparring with each other. The third person extends arms to stop them. The fourth person \textbf{engages in the fight}.}}
    {\setlength{\tabcolsep}{0pt}
    \begin{tabular}{*{8}{c}}
        \includegraphics[width=0.125\textwidth]{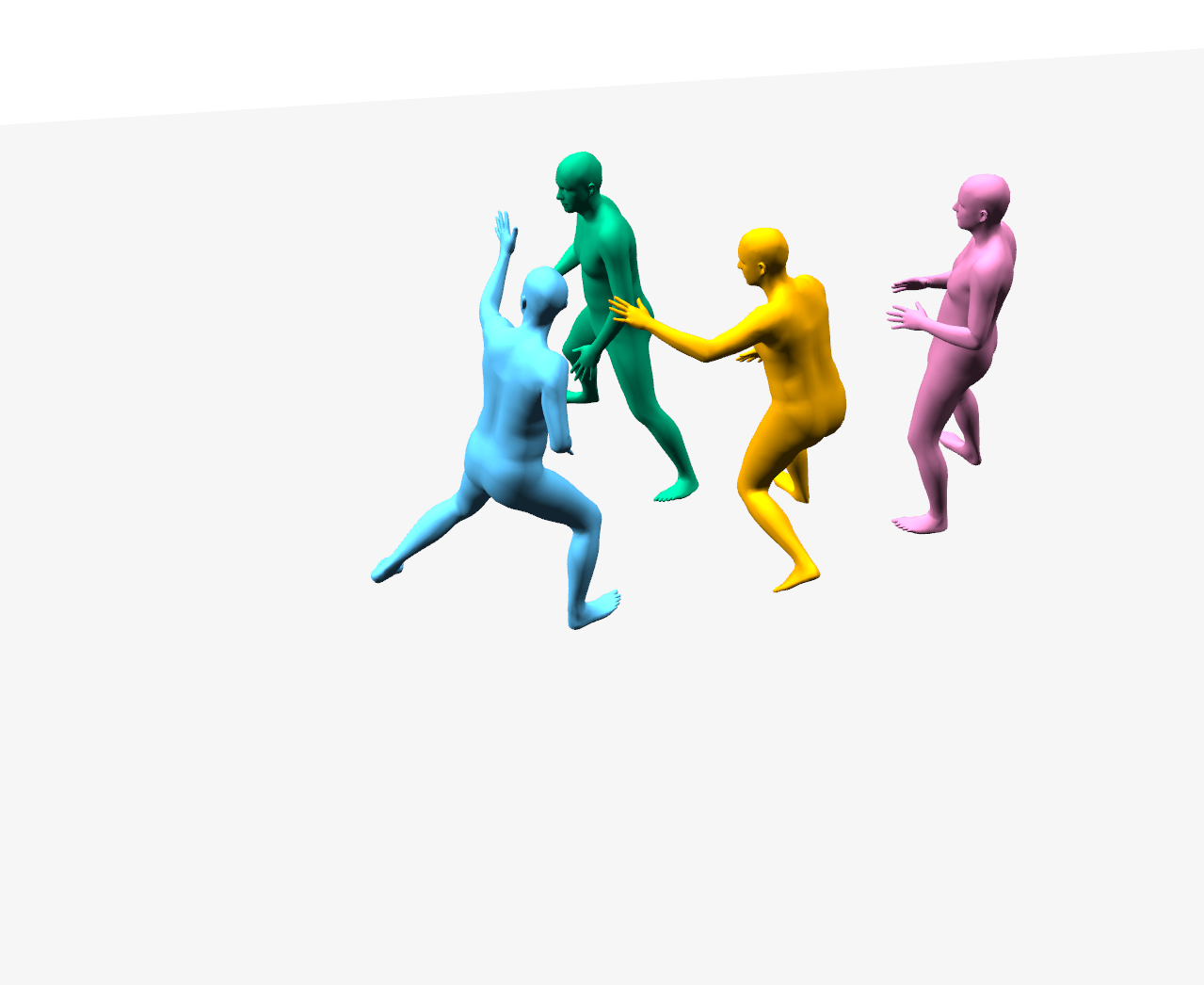} &
        \includegraphics[width=0.125\textwidth]{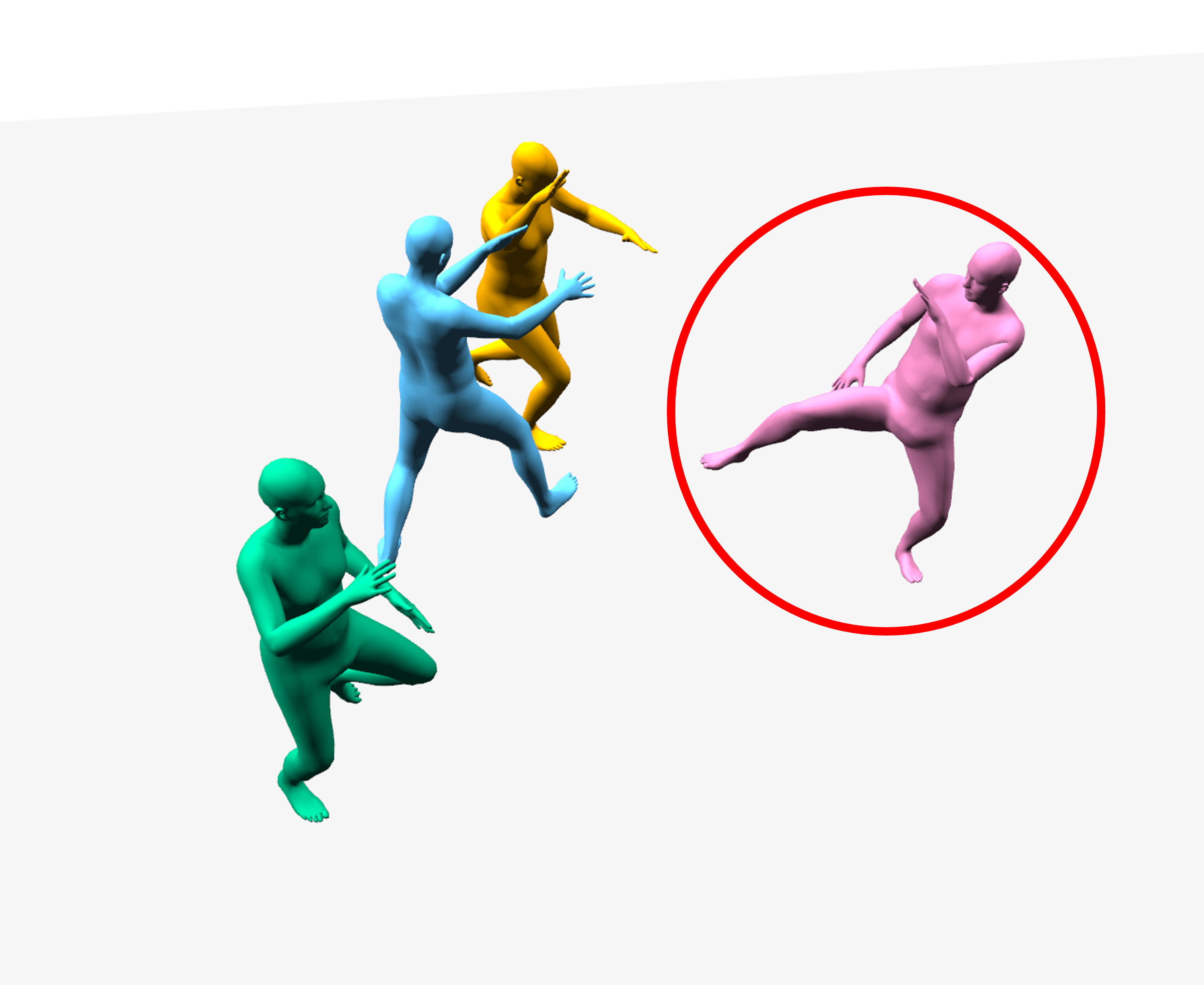} &
        \includegraphics[width=0.125\textwidth]{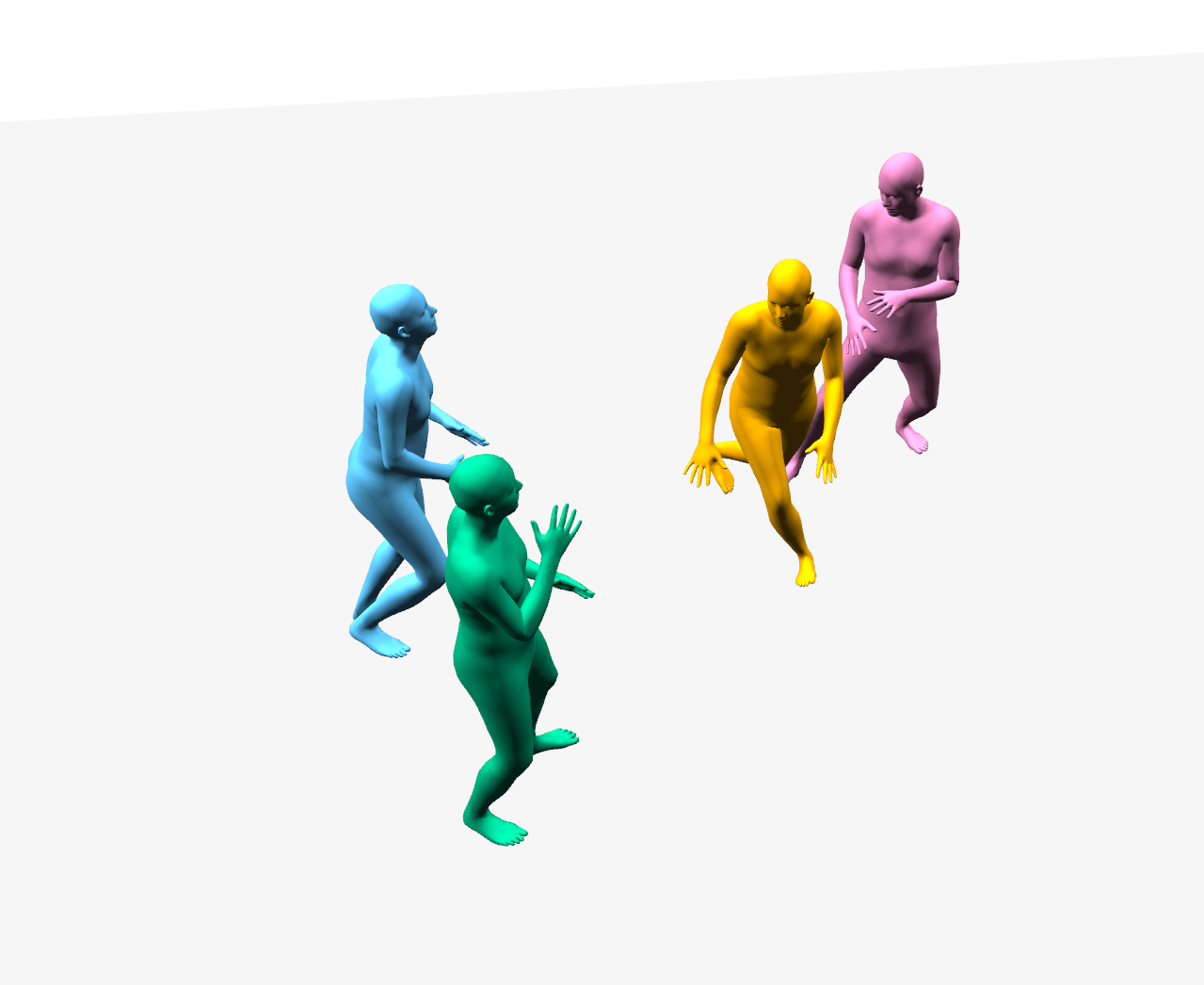} &
        \includegraphics[width=0.125\textwidth]{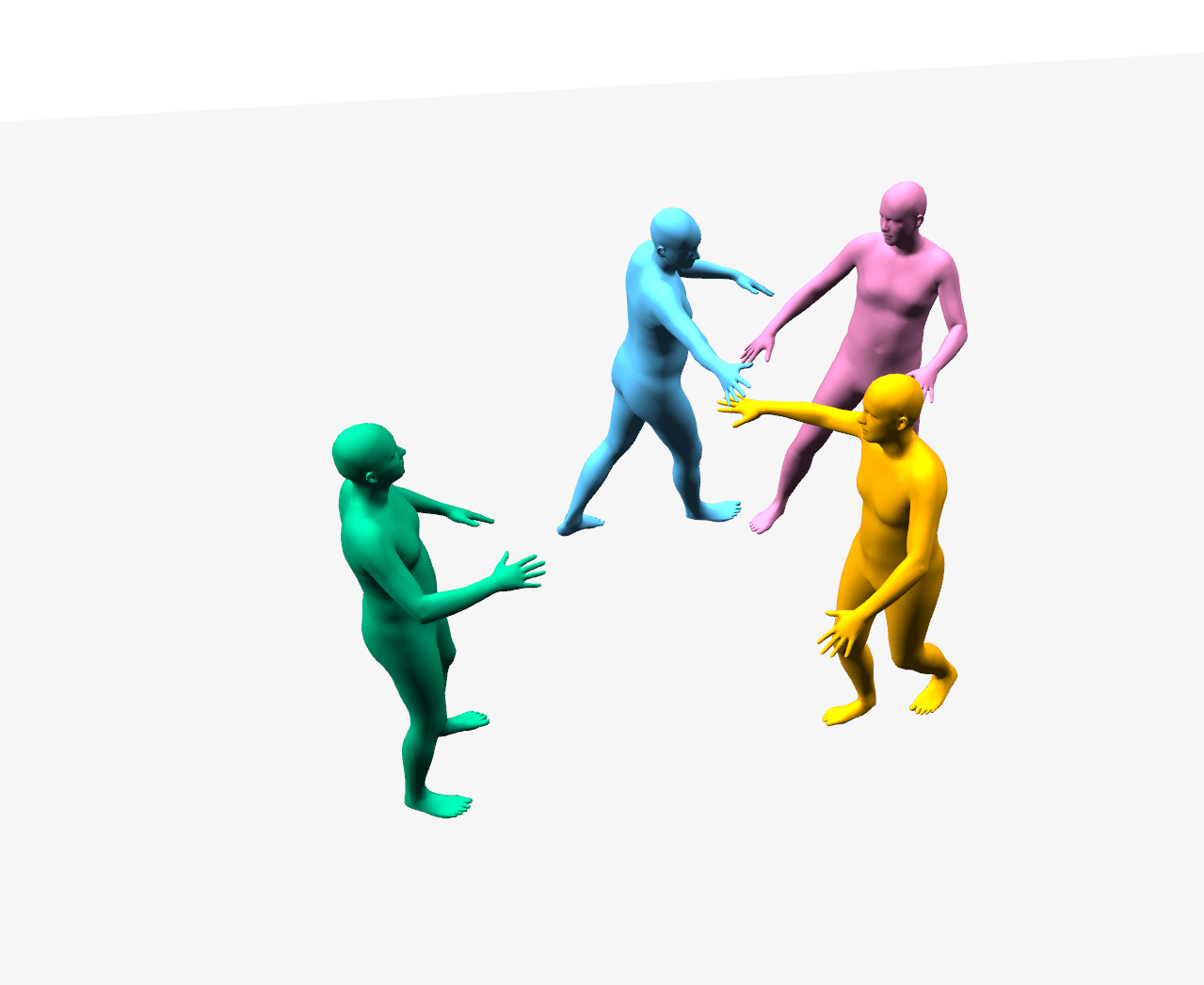} \hspace*{5pt}& 
        
        \includegraphics[width=0.125\textwidth]{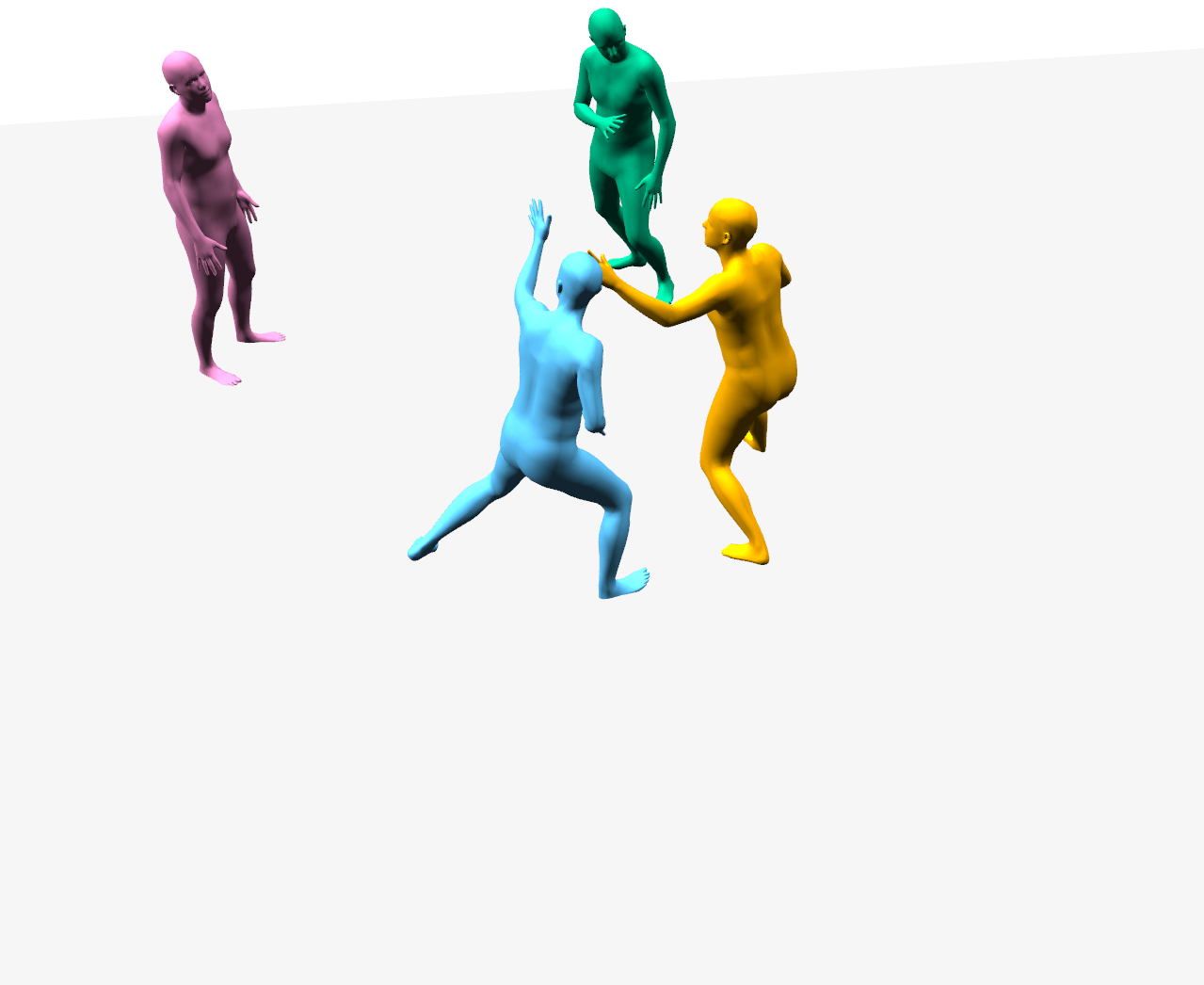} &
        \includegraphics[width=0.125\textwidth]{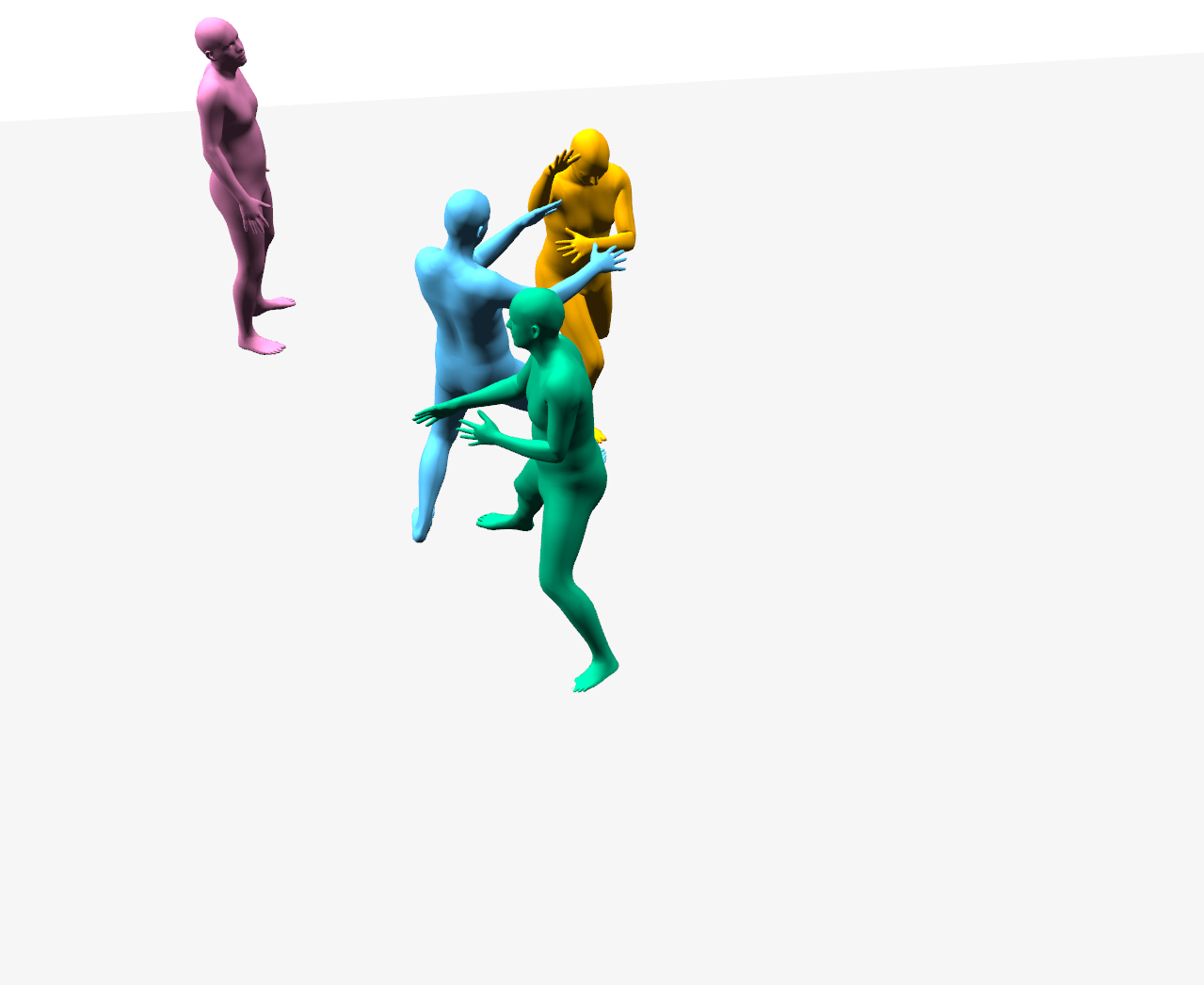} &
        \includegraphics[width=0.125\textwidth]{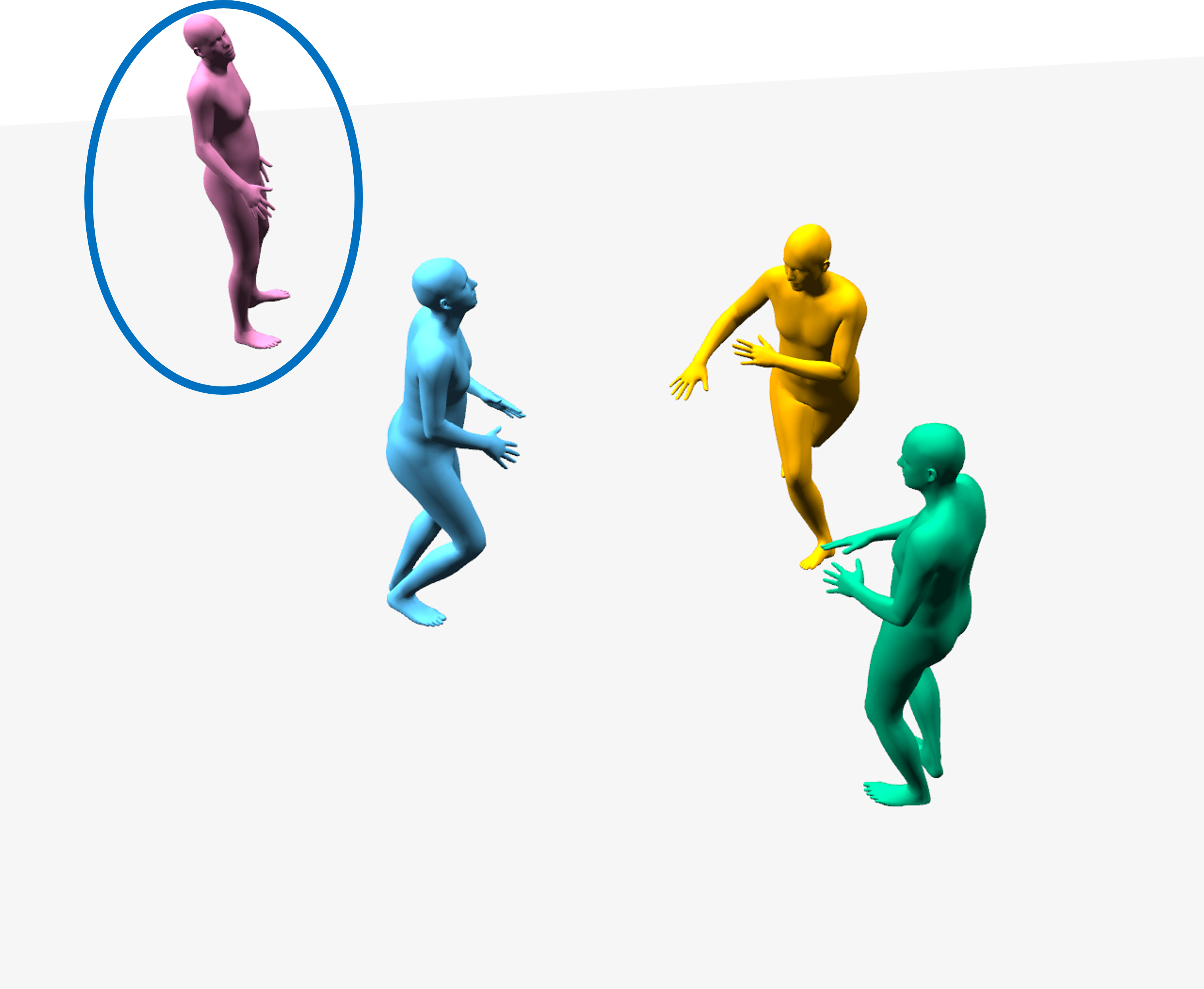} &
        \includegraphics[width=0.125\textwidth]{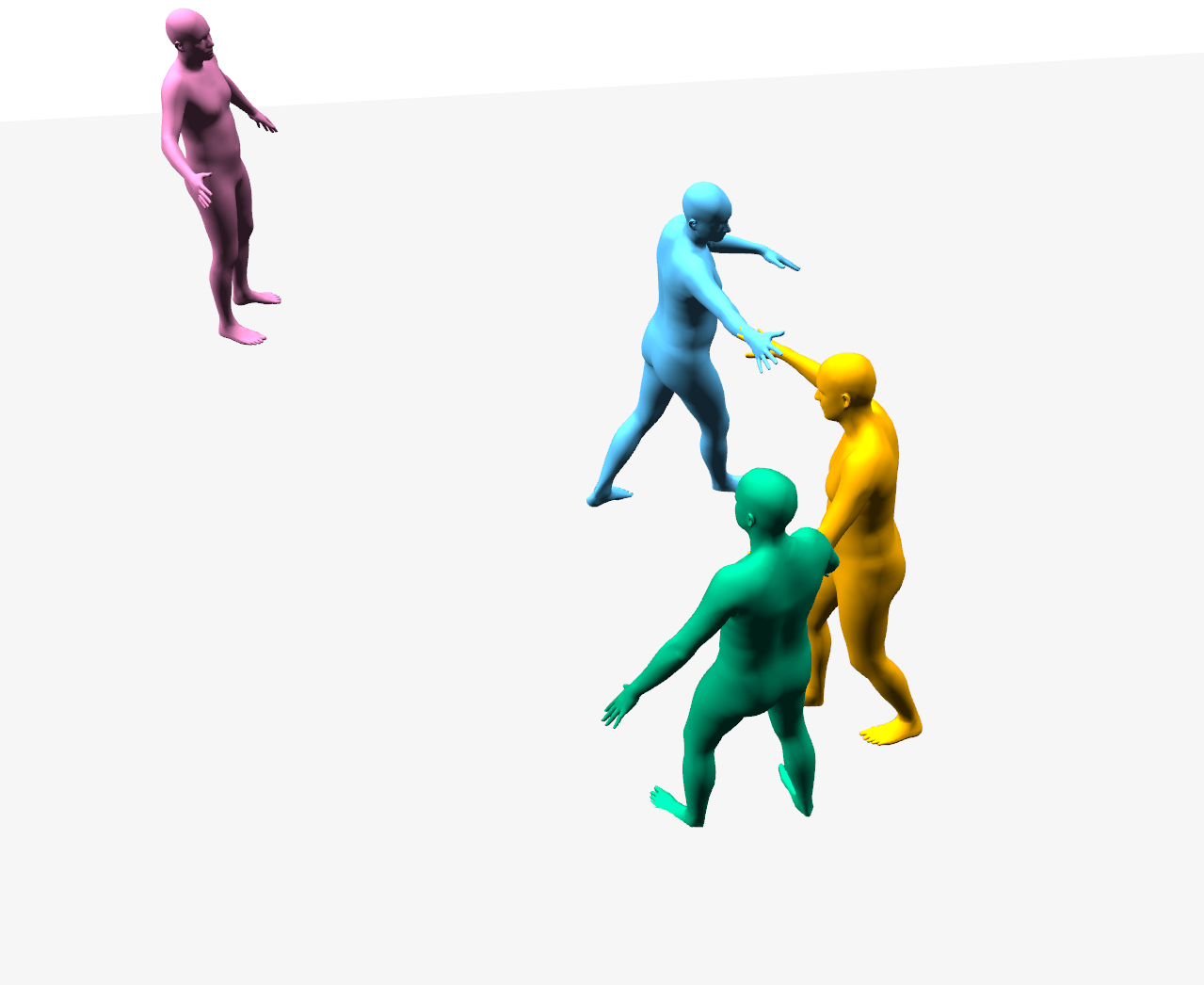} \\ 
    \end{tabular}}

    {\itshape \scriptsize{Five performers are \textbf{training in taekwondo} by exchanging attacks.}}
    
    {\setlength{\tabcolsep}{0pt}
    
    \begin{tabular}{*{8}{c}}
    
    
    \includegraphics[width=0.125\textwidth]{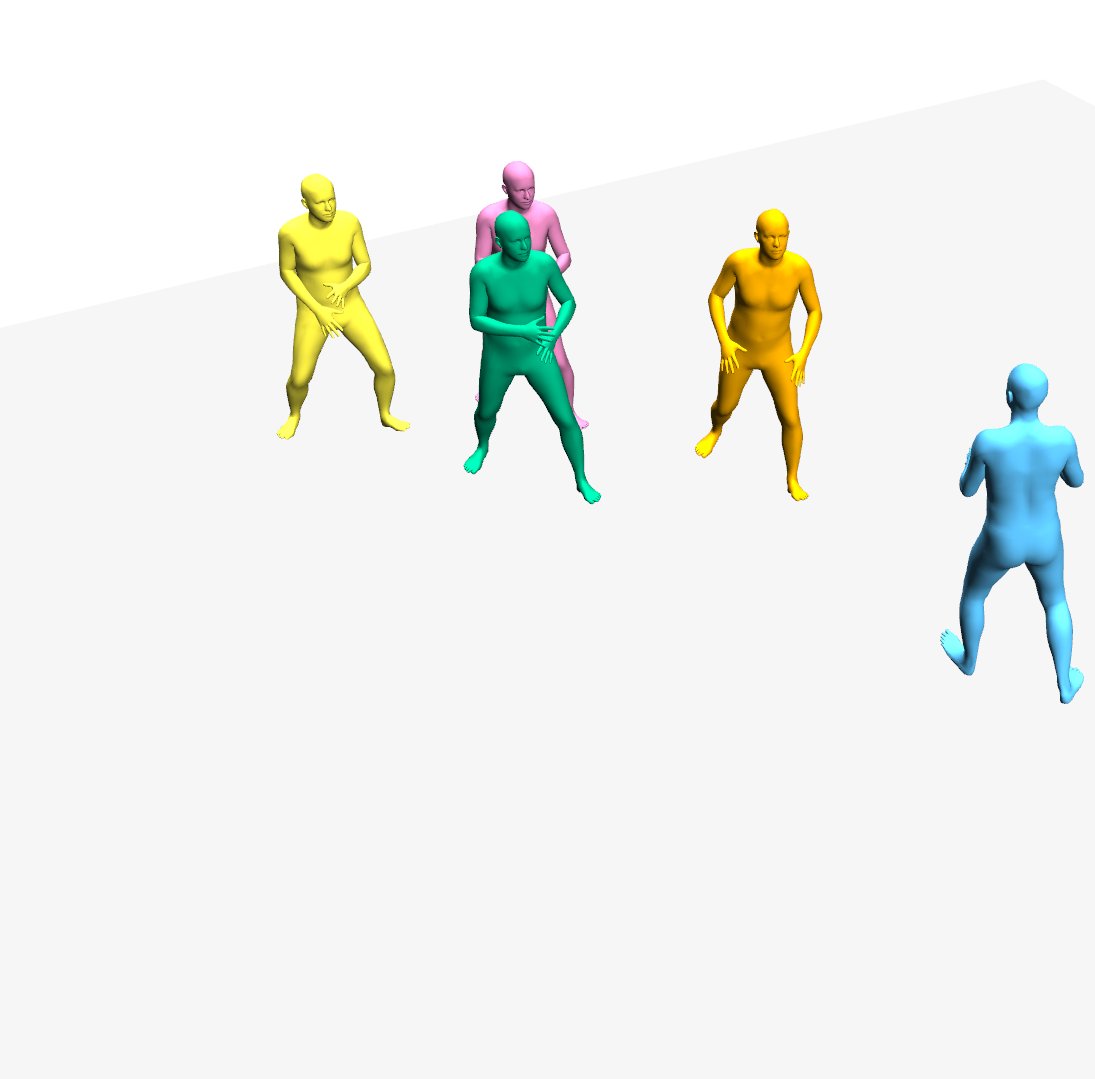} &
    \includegraphics[width=0.125\textwidth]{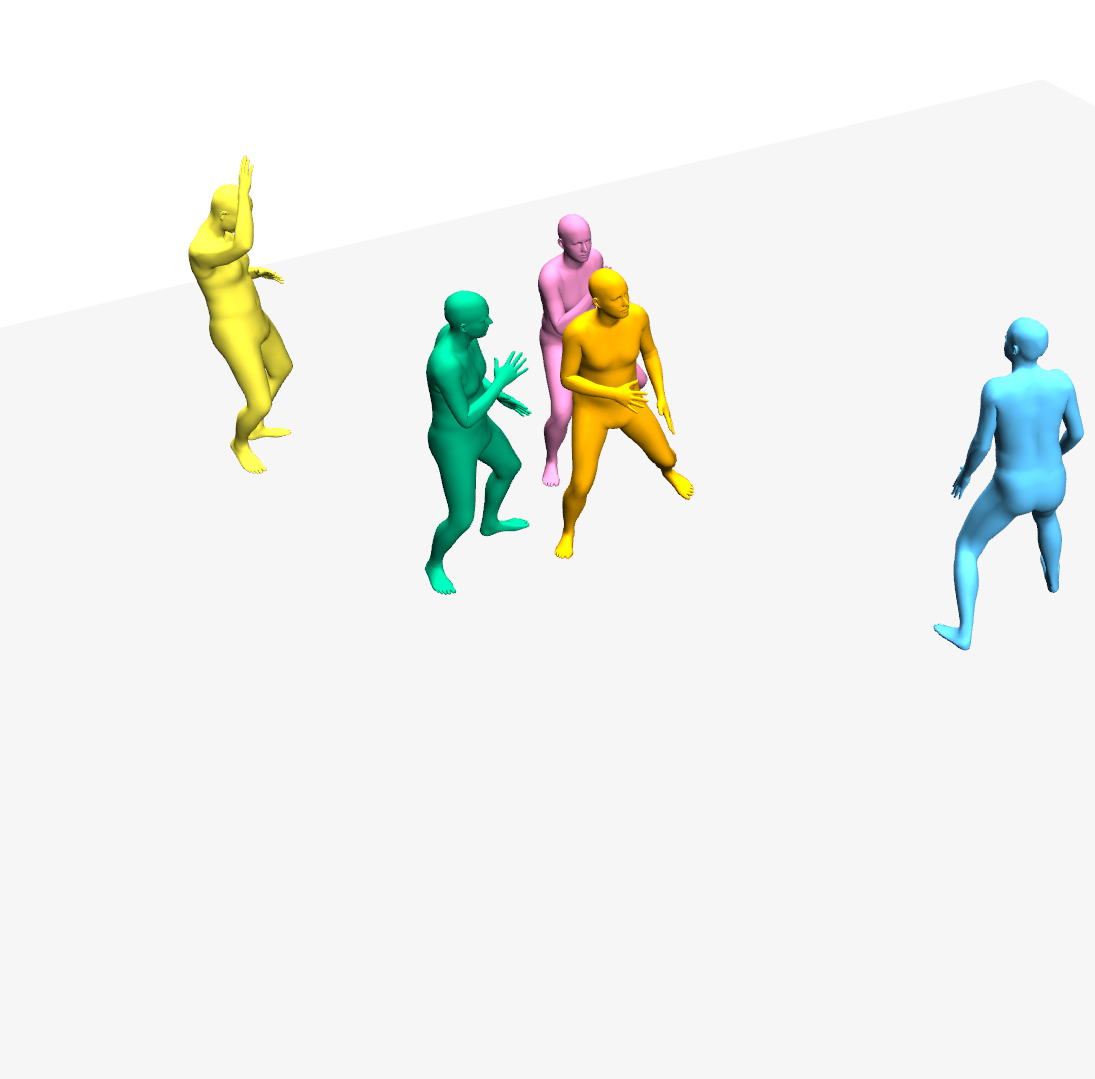} &
    \includegraphics[width=0.125\textwidth]{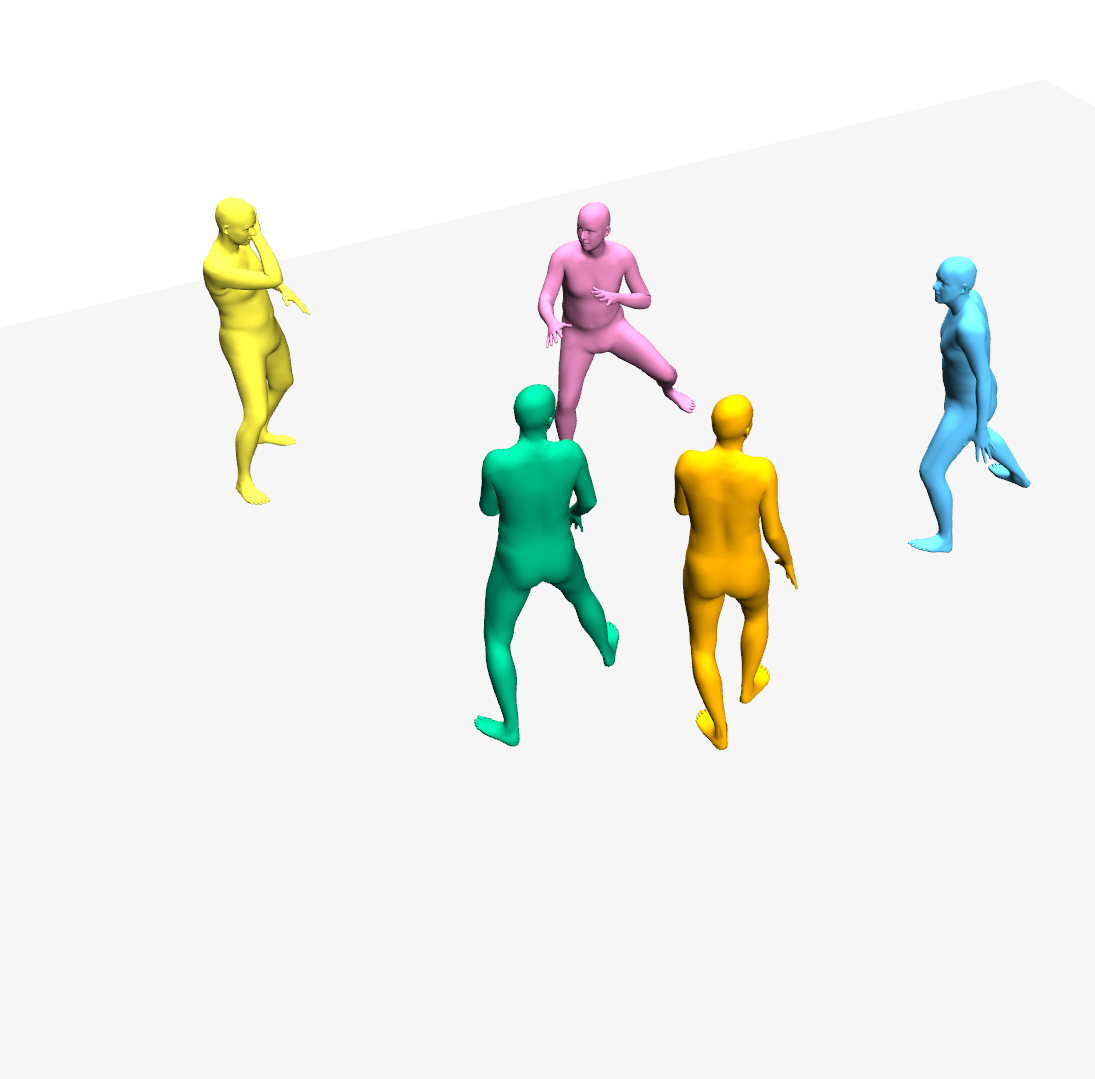} &
    \includegraphics[width=0.125\textwidth]{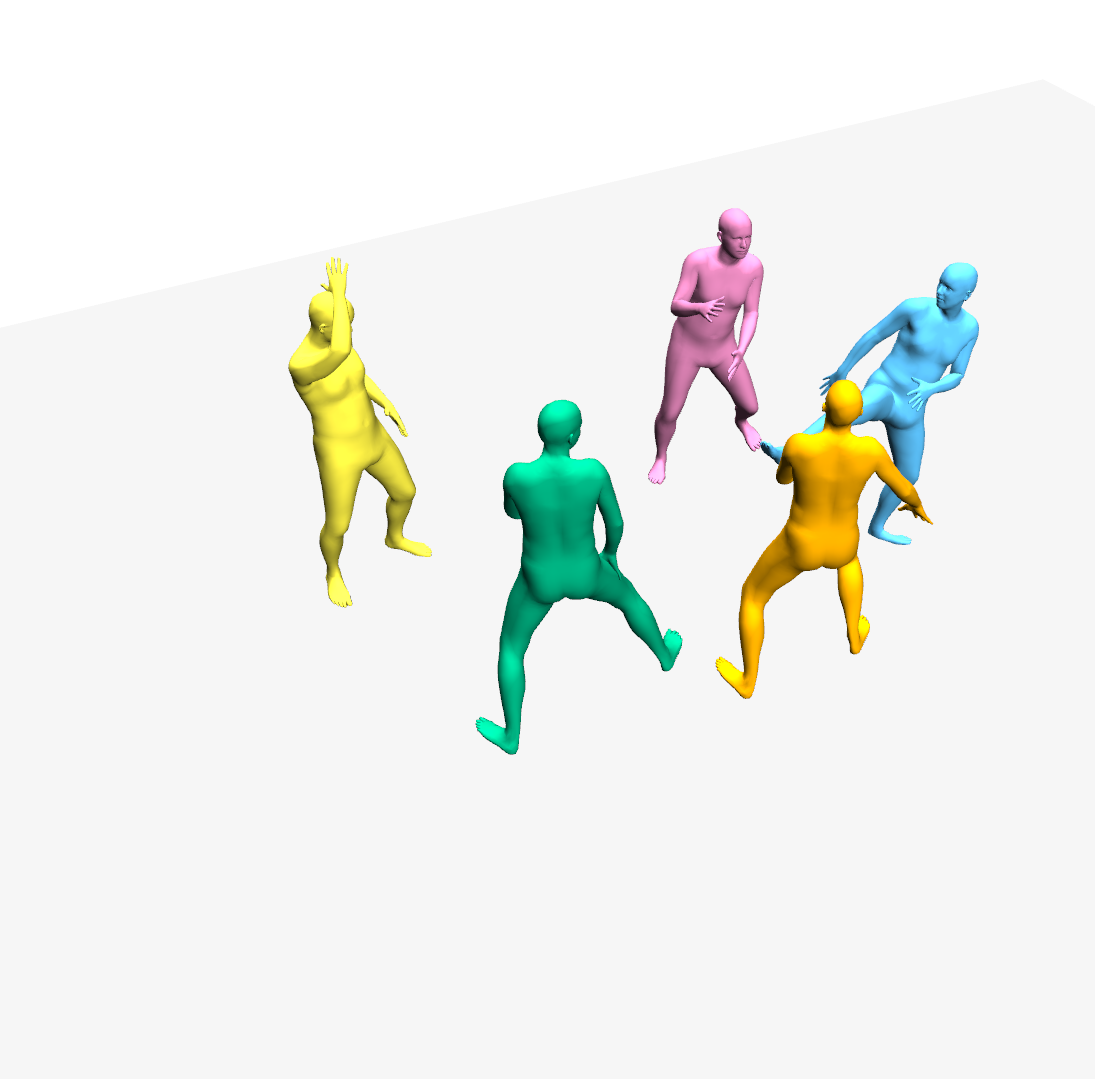} \hspace*{5pt}& 
    \includegraphics[width=0.125\textwidth]{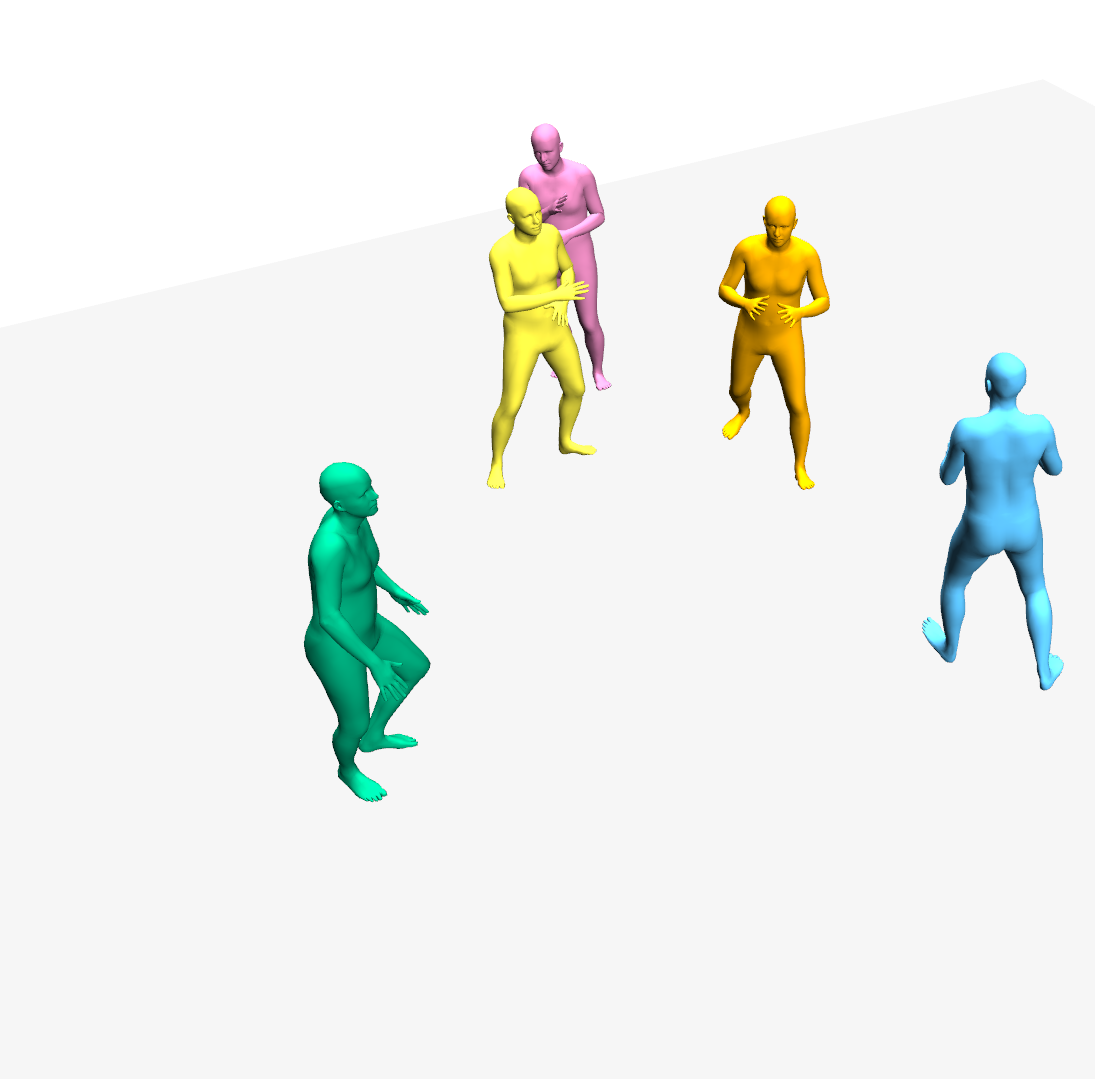} &
    \includegraphics[width=0.125\textwidth]{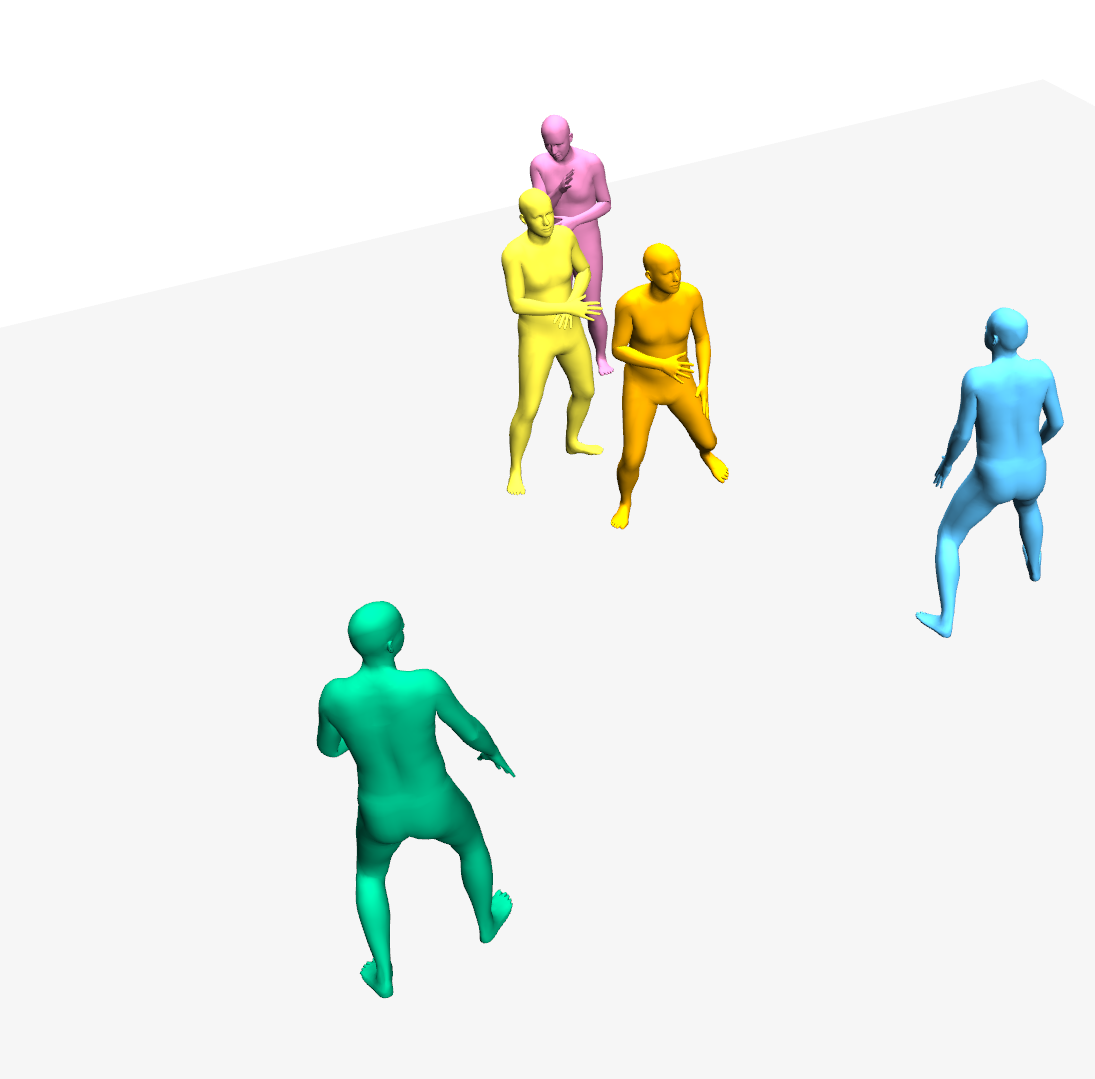} &
    \includegraphics[width=0.125\textwidth]{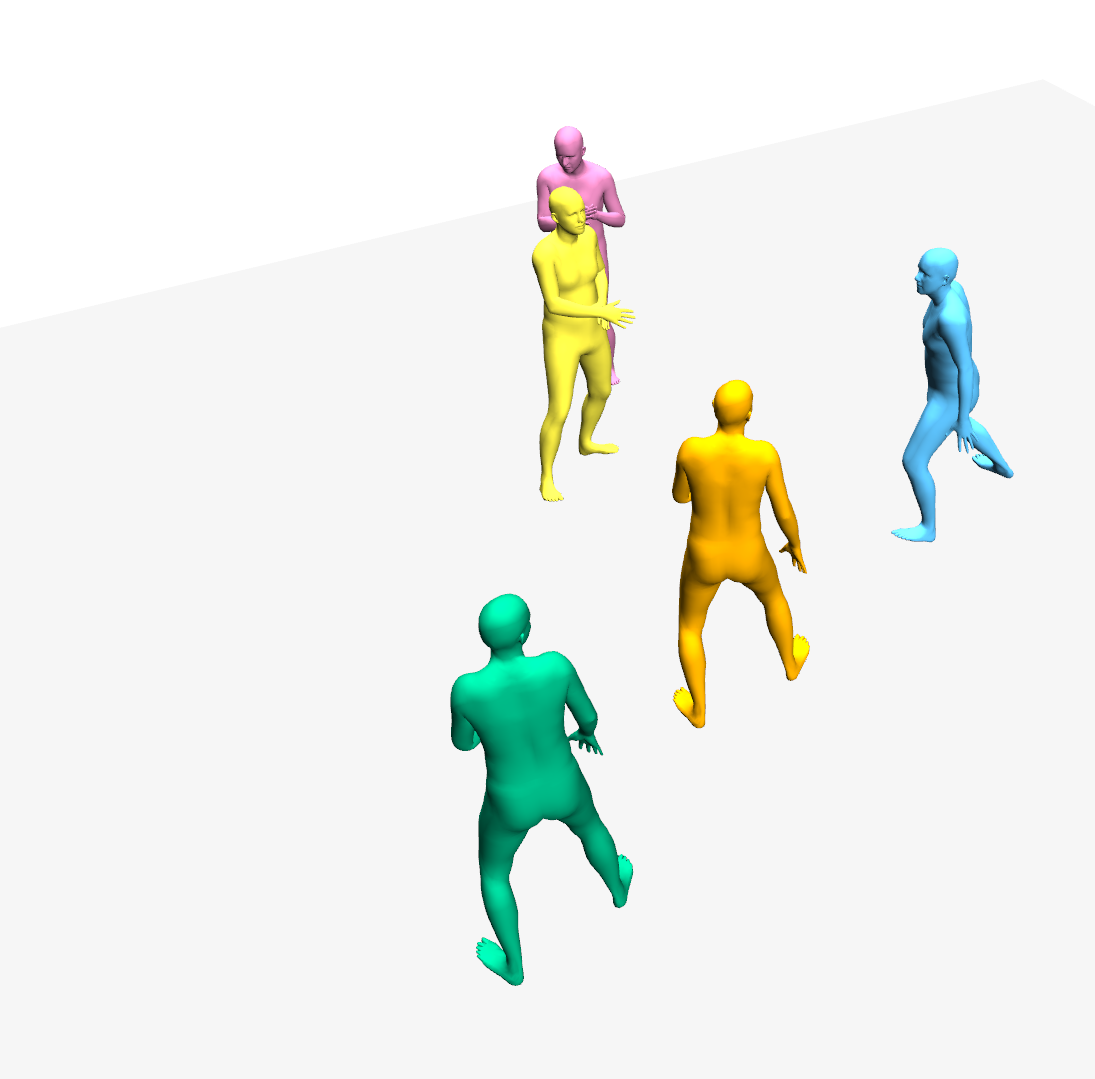} &
    \includegraphics[width=0.125\textwidth]{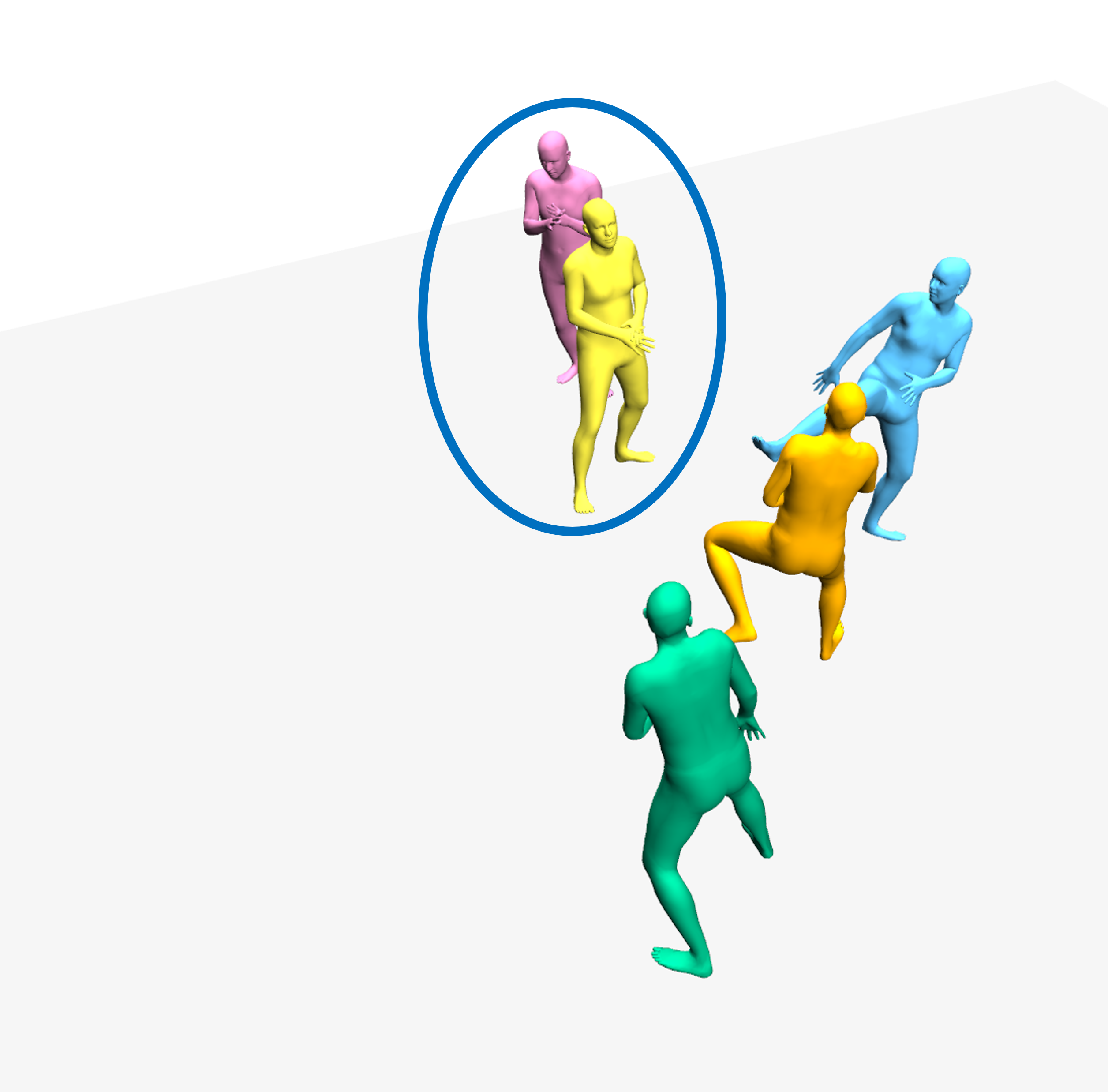} \\ 
    
    
    
    
    \end{tabular}}
    \vspace{-3mm}
    \caption{
        Qualitative comparison (zoom into see it better) between FreeMotion~\cite{fan2025freemotion} and UMF. {\color{red}Red circles} demonstrate successful cases, while {\color{blue}Blue circles} show failure cases.
    }
    \label{qual_fig}
    \vspace{-3mm}

\end{figure*}

\textbf{Adaptive Context Formulation.}
The adapter first produces the context motion $C_i$ by encoding the set of previously generated motions $\mathcal{Z}_{gen} $ with a transformer encoder: 
\small
\begin{align}
    C_{i} = TranEnc(\mathcal{Z}_{gen}).
\end{align}
\normalsize
Subsequently, if $i>2$, agent-wise average pooling is applied to match the latent dimension of $\hat{Z}_i$.
This design adaptively refines $\mathcal{Z}_{gen}$ into a concise global context, which alleviates error accumulation (See cases in Fig.~\ref{qual_fig}).

%

%
%

\textbf{S-Flow forward process.}
Similar to previous works~\cite{lipman2022flow, albergo2022building}, we use the rectified flow as the backbone, which is parameterized by a neural network $G^S_\theta$ to predict vector fields, i.e., $v = w_1 - w_0$.
%
%
S-Flow is trained by jointly modeling two probabilistic paths for reaction transformation and context reconstruction as follows: 

\noindent(1) For the reaction path, we interpolate between the previously generated motion (context) $w_0=C$ and the target reaction motion $w_1=W$, $w_t^\text{react}$ at timestep $t$ is: 
\small
\begin{equation}
w_t^\text{react} = tw_1 + [1 - t]w_0.
\label{eq_react_path}
\end{equation}
\normalsize
%
The training objective of the reaction transformation is:
\small
\begin{equation}
\mathcal{L}_\text{trans} = \mathbb{E}_{t,w_1, w_0} \|G^S_\theta(w_t^\text{react}, t, c) - (W-C)\|_2^2,
\label{eq_react_obj}
\end{equation}
\normalsize
where $c$ refers to the text prompt.

\noindent(2) For the context path, we interpolate between Gaussian noise $w'_0 =\epsilon$ and context motion $w'_1=C$, $w_t^\text{cont}$ at timestep $t$ is: 
\small
\begin{equation}
w_t^\text{cont} = tw'_1 + [1 - t]w'_0.
\label{eq_cont_path}
\end{equation}
\normalsize
The training objective of the context reconstruction is:
\small
\begin{equation}
\mathcal{L}_\text{recon} = \mathbb{E}_{t,w_0, \epsilon} \|G^S_\theta(w_t^\text{cont}, t, c) - (C - \epsilon)\|_2^2,
\label{eq_cont_obj}
\vspace{-2mm}
\end{equation}
\normalsize
where $c$ refers to the text prompt.

Finally, the S-Flow training objective is a weighted sum of these two losses $\mathcal{L}_\text{S-Flow} = \mathcal{L}_\text{trans} + \lambda_\text{recon} \mathcal{L}_\text{recon}.$
%
Thus $G^S_\theta$ learns to predict reaction for the next agent while being aware of the current context, balanced by $\lambda_\text{recon}$.

\textbf{S-Flow sampling process.}
As detailed in Algorithm 1, the sampling process mirrors P-Flow by using an Euler ODE solver. The discretization process involves dividing the procedure into $M=T_{S}$ steps, as follows:
\small
\begin{equation}
\hat{w}_{t_{m+1}} \leftarrow \hat{w}_{t_m} + (t_{m+1} - t_m) G^S_\theta(\hat{w}_{t_m}, t_m, c),
\label{odes}
\vspace{-2mm}
\end{equation}
\normalsize
where the integer time steps $t_1 = 0 < t_2 < \cdots < t_M = 1$.
The trajectory starts from the motion context $C$ from the context adapter layer, and ends with the reaction motion $W$.
%
%

\subsection{Justification of design choices}


\textbf{Asymmetric Inference Budget for UMF Efficiency.}
Generating motion for $N$ agents requires one P-Flow execution and $N-1$ S-Flow executions.
This structure motivates an asymmetric inference budget, as the quality of the motion prior determines the upper bound for all subsequent reactions.
We therefore allocate a substantial budget to P-Flow (\eg, 50 steps), which remains computationally feasible due to its pyramid structure.
We find the performance of P-Flow is sensitive to the total number of steps, but far less sensitive to the ratio of low-to-high resolution steps.
This allows us to assign more inference steps at low resolution (\eg, 45 steps), minimizing the overhead from the multi-token representation.
%
Furthermore, this dedicated motion prior enables the S-Flow to generate reactions with a minimal inference budget (\eg, 10 steps), keeping UMF computationally tractable when $N$ becomes large.

%


\noindent \textbf{Shared transformer between P-Flow and S-Flow.}
Sharing the transformer backbone between P-Flow and S-Flow would reduce the overall parameter count~\cite{fan2025freemotion}. However, we found that a shared backbone struggles to converge and yields degraded performance (See Tab.~\ref{abs3}).
We attribute this to two factors:
1) P-Flow focuses on mapping noise to motion, while S-Flow learns both motion-to-motion and noise-to-motion paths. These tasks are incompatible and challenging to optimize.
%
2)  The continuity guarantees~\cite{jin2024pyramidal} at the pyramid jump points are difficult to maintain, which assume tractable distributions (e.g., Gaussian noise), while S-Flow operates on complex motion distributions with intractable means and variances.
%
Therefore, UMF employs separate P-Flow and S-Flow modules.
The S-Flow transformer is shared autoregressively to generate the reaction for the subsequent agent, using all previously generated motions as context.

\begin{table*}[!h]
\caption{Quantitative evaluation on the InterHuman test sets. $\pm$ indicates a 95\% confidence interval and $\rightarrow$ means the closer to ground truth the better. \textbf{Boldface} indicates the best result, while \underline{underline} refers to the second best. 
}
\label{main_tab}
\centering
\scalebox{0.75}{
\begin{tabular}{lccccccc}
\toprule
Method & R Top 1$\uparrow$ & R Top 2$\uparrow$ & R Top 3$\uparrow$ & FID$\downarrow$ & MM Dist$\downarrow$ & Diversity$\rightarrow$ & MModality$\uparrow$ \\
\midrule
\midrule
Ground Truth & $0.452^{\pm0.008}$ & $0.610^{\pm0.009}$ & $0.701^{\pm0.008}$ & $0.273^{\pm0.007}$ & $3.755^{\pm0.008}$ & $7.948^{\pm0.064}$ & - \\
\cmidrule{1-8}
TEMOS~\cite{petrovich2022temos} & $0.224^{\pm0.010}$ & $0.316^{\pm0.013}$ & $0.450^{\pm0.018}$ & $17.375^{\pm0.043}$ & $6.342^{\pm0.015}$ & $6.939^{\pm0.071}$ & $0.535^{\pm0.014}$ \\
T2M~\cite{guo2022generating} & $0.238^{\pm0.012}$ & $0.325^{\pm0.010}$ & $0.464^{\pm0.014}$ & $13.769^{\pm0.072}$ & $5.731^{\pm0.013}$ & $7.046^{\pm0.022}$ & $1.387^{\pm0.076}$ \\
MDM~\cite{tevet2022human} & $0.153^{\pm0.012}$ & $0.260^{\pm0.009}$ & $0.339^{\pm0.012}$ & $9.167^{\pm0.056}$ & $7.125^{\pm0.018}$ & $7.602^{\pm0.045}$ & $\mathbf{2.350^{\pm0.080}}$ \\
ComMDM~\cite{shafir2023human} & $0.223^{\pm0.009}$ & $0.334^{\pm0.008}$ & $0.466^{\pm0.010}$ & $7.069^{\pm0.054}$ & $6.212^{\pm0.021}$ & $7.244^{\pm0.038}$ & $1.822^{\pm0.052}$ \\
InterGen~\cite{liang2024intergen} & $0.371^{\pm0.010}$ & $0.515^{\pm0.012}$ & $0.624^{\pm0.010}$ & $5.918^{\pm0.079}$ & $5.108^{\pm0.014}$ & $7.387^{\pm0.029}$ & \underline{$2.141^{\pm0.063}$} \\
MoMat-MoGen~\cite{cai2024digital} & $0.449^{\pm0.004}$ & $0.591^{\pm0.003}$ & $0.666^{\pm0.004}$ & $5.674^{\pm0.085}$ & $3.790^{\pm0.001}$ & $8.021^{\pm0.035}$ & $1.295^{\pm0.023}$ \\
in2IN~\cite{ponce2024in2in} & $0.425^{\pm0.008}$ & $0.576^{\pm0.008}$ & $0.662^{\pm0.009}$ & $5.535^{\pm0.120}$ & $3.803^{\pm0.002}$ & \underline{$7.953^{\pm0.047}$} & $1.215^{\pm0.023}$ \\
TIMotion~\cite{wang2025timotion} & $\mathbf{0.491^{\pm0.005}}$ & $\mathbf{0.648^{\pm0.004}}$ & $\mathbf{0.724^{\pm0.004}}$ & $5.433^{\pm0.080}$ & $\mathbf{3.775^{\pm0.001}}$ & $8.032^{\pm0.030}$ & $0.952^{\pm0.032}$ \\
InterMask~\cite{javed2024intermask} & $0.449^{\pm0.004}$ & $0.599^{\pm0.005}$ & $0.683^{\pm0.004}$ & $\underline{5.154^{\pm0.061}}$ & $3.790^{\pm0.002}$ & $\mathbf{7.944^{\pm0.033}}$ & $1.737^{\pm0.020}$ \\
\cmidrule{1-8}
FreeMotion & $0.326^{\pm0.003}$ & $0.462^{\pm0.006}$ & $0.544^{\pm0.006}$& $6.740^{\pm0.130}$ & $3.848^{\pm0.002}$ & $7.828^{\pm0.130}$ & $1.226^{\pm0.046}$ \\
UMF & \underline{$0.467^{\pm0.004}$} & \underline{$0.620^{\pm0.004}$} & \underline{$0.694^{\pm0.005}$} & $\mathbf{4.772^{\pm0.079}}$ & \underline{$3.784^{\pm0.001}$} & $8.039^{\pm0.032}$ & $1.398^{\pm0.012}$ \\
\midrule
\end{tabular}
}
\vspace{-6mm}
\end{table*}


\begin{table}[!h]
    \centering
    \caption{Comparison to state-of-the-art for human action-reaction synthesis on the InterHuman-AS dataset. $\pm$ indicates 95\% confidence interval, $\rightarrow$ means that closer to Real is better. \textbf{Bold} indicates best result and \underline{underline} indicates second best.}
    \label{tab:comparison}
    \vspace{-2mm}
    \resizebox{0.48\textwidth}{!}{
    \begin{tabular}{lccccc}
    \toprule
    Methods & RTop3$\uparrow$ & FID $\downarrow$ & MM Dist$\downarrow$ & Diversity$\rightarrow$ & MModality $\uparrow$ \\
    \midrule
    Real & $0.722^{\pm 0.004}$ & $0.002^{\pm 0.0002}$ & $3.503^{\pm 0.011}$ & $5.390^{\pm 0.058}$ & - \\
    \midrule
    T2M~\cite{guo2022generating} & $0.224^{\pm 0.003}$ & $32.482^{\pm 0.098}$ & $7.299^{\pm 0.016}$ & $4.350^{\pm 0.073}$ & $0.719^{\pm 0.041}$ \\
    MDM~\cite{tevet2022human} & $0.370^{\pm 0.006}$ & $3.397^{\pm 0.052}$ & $8.640^{\pm 0.045}$ & $4.780^{\pm 0.015}$ & $2.288^{\pm 0.039}$ \\
    MDM-GRU~\cite{tevet2022human} & $0.328^{\pm 0.012}$ & $6.397^{\pm 0.214}$ & $8.884^{\pm 0.040}$ & \underline{$4.851^{\pm 0.081}$} & $2.076^{\pm 0.040}$ \\
    RAIG~\cite{tanaka2023role} & $0.363^{\pm 0.008}$ & $2.915^{\pm 0.029}$ & $7.294^{\pm 0.027}$ & $4.736^{\pm 0.099}$ & $2.203^{\pm 0.049}$ \\
    InterGen~\cite{liang2024intergen} & $0.374^{\pm 0.005}$ & $13.237^{\pm 0.035}$ & $10.929^{\pm 0.026}$ & $4.376^{\pm 0.042}$ & $\mathbf{2.793^{\pm 0.014}}$ \\
    ReGenNet~\cite{xu2024regennet} & $0.407^{\pm 0.003}$ & $\mathbf{2.265^{\pm 0.097}}$ & \underline{$6.860^{\pm 0.040}$} & $\mathbf{5.214^{\pm 0.139}}$ &$2.391^{\pm 0.023}$ \\
    \midrule
    FreeMotion~\cite{fan2025freemotion} & \underline{$0.409^{\pm 0.006}$} & $3.896^{\pm 0.029}$ & $7.632^{\pm 0.036}$ & $6.089^{\pm 0.027}$ & \underline{$2.496^{\pm 0.036}$} \\
    UMF & $\mathbf{0.530^{\pm 0.006}}$ & \underline{$2.577^{\pm 0.024}$} & $\mathbf{4.987^{\pm 0.011}}$ & $7.764^{\pm 0.024}$ & $2.116^{\pm 0.041}$ \\
    \bottomrule
    \end{tabular}}
    \vspace{-8mm}
\end{table}

\vspace{-2mm}
\begin{table}[h]
    \centering
    \caption{Ablation study of individual priors on the HumanML3D and InterHuman datasets. \textbf{HP}: \textbf{H}eterogeneous \textbf{P}riors; \textbf{LA}: \textbf{L}atent \textbf{A}dapter; \textbf{MT}: \textbf{M}ulti-token \textbf{T}okenizer.}
    \vspace{-2mm}
    \resizebox{0.48\textwidth}{!}{
    \begin{tabular}{l|ccc|cc|cc}
        \hline
        & \multicolumn{3}{c|}{Components} & \multicolumn{2}{c|}{HumanML3D} & \multicolumn{2}{c}{InterHuman} \\
        \hline
        Method & HP & LA & MT & RTop3 $\uparrow$ & FID $\downarrow$ & RTop3$\uparrow$ & FID$\downarrow$ \\
        \hline
        Real Data & - & - & - & 0.797 & 0.002 & 0.701 & 0.273 \\
        \hline\hline
        FreeMotion & \checkmark & - & - & 0.612 & 3.539 & 0.503 & 7.984 \\
        \textit{w/o HP} & \ding{55} & - & - & 0.128 & 13.155 & 0.544 & 6.740 \\
        \hline
        UMF (Full) & \checkmark & \checkmark & \checkmark & 0.729 & 0.486 & 0.694 & 4.772 \\
        \textit{w/o HP} & \ding{55} & \checkmark & \checkmark & 0.134 & 10.806 & 0.651 & 4.933 \\
        \textit{w/o LA} & \checkmark & \ding{55} & \checkmark & 0.708 & 0.646 & 0.627 & 5.473 \\
        \textit{w/o MT} & \checkmark & \checkmark & \ding{55} & 0.713 & 0.534 & 0.655 & 5.231 \\
        \textit{w/o (HP + LA)} & \ding{55} & \ding{55} & \checkmark & 0.131 & 12.914 & 0.579 & 5.561 \\
        \textit{w/o (HP + MT)} & \ding{55} & \checkmark & \ding{55} & 0.139 & 12.748 & 0.616 & 5.493 \\
        \textit{w/o (LA + MT)} & \checkmark & \ding{55} & \ding{55} & 0.682 & 0.845 & 0.605 & 5.668  \\
        \textit{w/o (HP + LA + MT)} & \ding{55} & \ding{55} & \ding{55} & 0.125 & 13.844 & 0.511 & 6.036 \\
        \hline
    \end{tabular}}
    \label{abs1}

\end{table}


\begin{table}[t]
\centering
 \caption{Ablation study of Pyramid Flow on the InterHuman dataset. UMF has a 2-stage temporal pyramid structure. We report FLOPs(G) and AITS (Average Inference Time in Seconds).
 $T_{P2}$, $T_{P1}$, and $T_{S}$ refer to the corresponding inference step for P-Flow low-res stage, P-Flow full-res stage, and S-Flow, respectively.
 UMF-PFK1 and UMF-PFS refer to P-Flow with the original resolution and with the spatial pyramid structure, respectively. 
 }
 \vspace{-2mm}
 \scalebox{0.7}{
 \begin{tabular}{lccccccc}
 \hline
Methods & $T_{P_2}$ & $T_{P_1}$ & $T_{S}$ & FID $\downarrow$ & RTop3$\uparrow$ & FLOPs(G)$\downarrow$ & AITS$\downarrow$ \\
 \hline
 FreeMotion & - & - & - & 6.740 & 0.544 & 217.8 &3.059 \\
 UMF & 45 & 5 & 10 & 4.772 & 0.694 & 140.3 & 0.623 \\
 \hline
 UMF-PFK1 & - & 50 & 10 & 4.761 & 0.674 & 320.2 & 1.119 \\
 UMF-PFS & 45 & 5 & 10 & 7.238 & 0.527 & 135.9 & 0.581 \\
 \hline
 UMF-Fast & 5 & 5 & 10 & 4.937 & 0.687 & 74.7 & 0.439 \\
 UMF-Symmetric & 25 & 25 & 10 & 4.784 & 0.697 & 206.0 & 0.874 \\
 \hline
\end{tabular}}
\label{abs2}
\vspace{-2mm}
\end{table}

\section{Experiments}

\subsection{Datasets, Metrics \& Implementation Details}
We utilize InterHuman~\cite{liang2024intergen} and HumanML3D~\cite{guo2022generating} datasets for the evaluation of text-conditioned motion generation performance.
The InterHuman and HumanML3D datasets contain 7,779 interaction sequences and 14,616 individual sequences, respectively, where each sequence is illustrated with 3 textual annotations.
The InterHuman-AS dataset~\cite{xu2024regennet} is essentially the same as InterHuman, but includes additional actor–reactor order annotations.
We employ the evaluation metrics following previous studies~\cite{liang2024intergen, guo2022generating}. Fidelity is assessed using Frechet Inception Distance (FID), R-precision, and Multimodal Distance (MM Dist), and diversity is evaluated with Diversity and Multimodality scores. 
%
All our models are trained with the AdamW optimizer using an initial learning rate of $10^{-4}$ and a cosine decay schedule.
Our mini-batch size is set to 128 during the VAE training stage and 64 during the flow matching training stage.
Each model was trained for 6K epochs during the VAE stage, 2K epochs during the P-Flow and 2K epochs during S-Flow stage. See Appendix C for details.

\subsection{Quantitative Results}
%
As shown in Table~\ref{main_tab}, on the InterHuman benchmark, UMF substantially outperforms the generalist baseline, FreeMotion~\cite{fan2025freemotion}, improving Top3 R-Precision by 28\% and reducing FID by 29\%. Furthermore, its Diversity score closely matches the ground truth, indicating a highly realistic output.
Against specialist methods tailored for dual-agent scenarios, UMF demonstrates competitive performance, outperforming the strongest baseline, InterMask~\cite{javed2024intermask}, by 7\% in FID.
It also achieves the second-best results on R-Precision and MM-Distance, demonstrating competitive text-following ability.
%
In Table~\ref{tab:comparison}, we compare UMF with existing approaches on the InterHuman-AS dataset, where we observe a similar trend. Specifically, UMF improves Top3 R-Precision by over 30\% and reduces MM-Distance by 27\% compared to ReGenNet~\cite{xu2024regennet}, significantly improving the reactive motion quality.


\subsection{Qualitative Results \& User Study}
\label{qr_us}
Fig.~\ref{qual_fig} demonstrates UMF's ability to generate more realistic human interactions compared to FreeMotion~\cite{fan2025freemotion}.
In the ``kick" (dual-agent) scenario, UMF generates a plausible kicking motion with correct leg assignment. In contrast, FreeMotion fails to produce a coherent generation, only attempting a poorly directed kick in the end.
In the "stroll" (three-agent) scenario, UMF correctly positions the third agent (green) between the other two (yellow, blue), maintaining plausible proximity, while FreeMotion's output suffers from severe interpenetration.
In the complex, multi-agent ($N>3$) ``fight" scenario, FreeMotion fails to animate all participants, resulting in artifacts such as the static poses of agents.
In contrast, UMF generalizes effectively to this zero-shot number-free task, producing dynamic and plausible interactions.

\begin{figure}[t] 
\centering    
\includegraphics[width=0.4\textwidth]{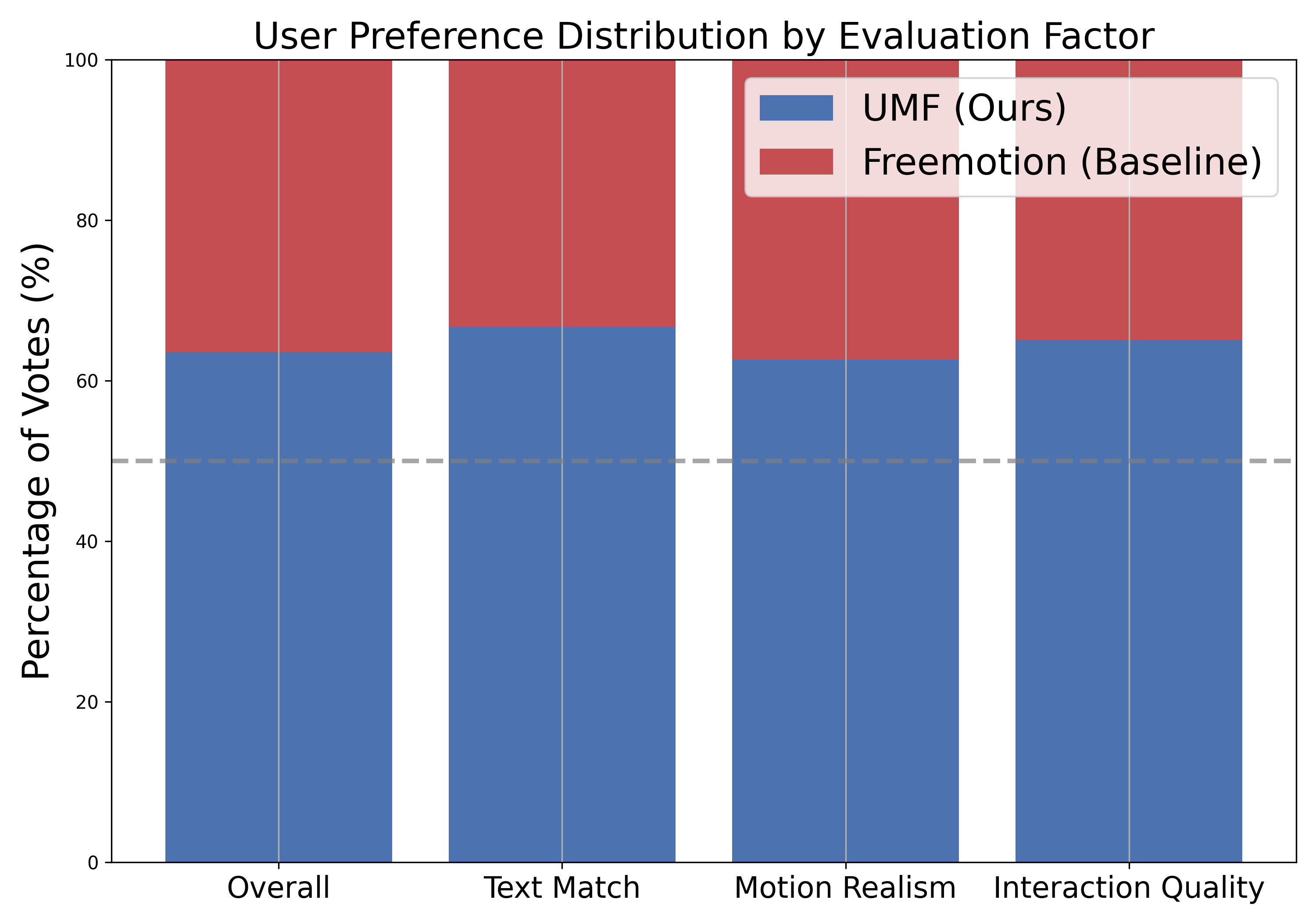}
\vspace{-3mm}
\caption{
The UMF number-free zero-shot generation user study. We asked users to compare our UMF ({\color{blue}Blue Bar}) to the FreeMotion ({\color{red}Red Bar}) in a side-by-side view. The dashed line marks 50\%. UMF outperforms FreeMotion in all three aspects of generation.
}
\vspace{-5mm}
\label{userstudy}
\end{figure}

Due to the scarcity of motion databases for group scenarios, we conducted a user study for assessing UMF's zero-shot generalization capability (see Appendix D).
The proposed UMF and FreeMotion~\cite{fan2025freemotion} are compared according to the aspects of text alignment, physical realism, interaction quality, and overall quality. 
30 unique users participated in the user study, with 20 randomly sampled multi-person generations ($N > 2$).
The zero-shot results in Fig.~\ref{userstudy} show that the number-free motions generated by UMF were clearly preferred over those generated by FreeMotion.



\vspace{-1mm}
\subsection{Ablation Studies}
\vspace{-1mm}

\noindent\textbf{Heterogeneous Priors and Latent Space.}
Table~\ref{abs1} investigates the impact of individual priors from HumanML3D~\cite{guo2022generating} and our latent space design. 
The results demonstrate that models trained with the HumanML3D prior outperform those without, improving both text adherence and motion fidelity. 
This highlights the potential of leveraging single-agent datasets to enhance multi-agent interaction generation.
We attribute the modest improvement to the complexity gap between the single-agent and multi-agent generation targets, which manifests as a challenging cross-dataset transfer effect.
Furthermore, we compare UMF against variants without the Latent Adapter (w/o. LA) and with a single-token latent space ($1 \times 256$). 
The results indicate that the Latent Adapter is crucial for multi-token flow matching, whereas the single-token variant lacks sufficient capacity to model number-free generation effectively.

\noindent\textbf{Efficiency Analysis of Pyramid Flow.}
Table~\ref{abs2} ablates the Pyramid Flow (PF) structure and its inference step allocation. 
First, we compare UMF with FreeMotion under the same inference steps (\ie, 60 steps), where UMF achieves lower FLOPs and is nearly 5$\times$ faster, which demonstrates the efficiency of P-Flow for complex interaction generation. 
Next, we compare two variants, where UMF-PFK1 achieves a slightly better FID but with extra computational cost. Conversely, the UMF-PFS variant shows severe performance degradation. 
We also find that reducing the P-Flow budget from 50 to 10 steps can nearly halve the FLOPs but degrade performance. 
Notably, allocating asymmetric steps ($T_{P2}$ = 45, $T_{P1}$ = 5) achieves the best speed-quality trade-off, yielding competitive FID with fewer FLOPs compared to a symmetric allocation ($T_{P2}$ = $T_{P1}$ = 25).

\noindent\textbf{Semi-Noise Flow Component Analysis.}
Table~\ref{abs3} analyzes the key components of S-Flow.
%
%
Sharing the transformer backbone between S-Flow and P-Flow, while parameter-efficient, results in significantly worse fidelity, likely due to their incompatible learning paths. 
We also compare the semi-noise flow with noise-free flow in~\cite{jiang2025arflow}, which only learns the reaction transformation path. This variant shows degraded performance without considering the error accumulation.
Removing the reconstruction loss also harms generation quality. Similarly, removing the Context Adapter entirely or replacing it with a ControlNet~\cite{zhang2023adding} both lead to a significant performance drop, underscoring the importance of context reconstruction. 
In contrast, the transformer-based adapter in UMF preserves a global view of the entire context.

\begin{table}[!t]
    \centering
    \caption{Ablation study of Semi-Noise Flow on the InterHuman dataset.}
    \vspace{-2mm}
    \label{tab:n-flow}
    \scalebox{0.8}{
    \begin{tabular}{lccc}
    \hline
    Methods  & FID $\downarrow$  & RTop3$\uparrow$  &Diversity$\rightarrow$\\
    \hline
    UMF  & 4.772 & 0.694  & 8.039 \\
    \hline
    UMF w. Shared Transformer & 6.206 & 0.644 & 8.088 \\
    UMF w. Noise-Free path~\cite{jiang2025arflow} & 5.617   & 0.646 & 8.112 \\
    UMF w. ControlNet~\cite{zhang2023adding} & 6.868 & 0.637 & 8.061 \\
    UMF w/o. Context Adapter & 7.038 & 0.642 & 8.087 \\ 
    UMF w/o. $L_{recons}$ & 5.765  & 0.649 & 8.124 \\
    \hline
    \end{tabular}}
    \vspace{-3mm}
\label{abs3}
\end{table}

\vspace{-3mm}

\section{Conclusion}
\vspace{-1mm}
We introduce Unified Motion Flow (UMF), a generalist framework for number-free, text-conditioned motion generation, which consists of Pyramid Motion Flow (P-Flow) and Semi-Noise Motion Flow (S-Flow).
%
Based on a unified heterogeneous latent space, UMF achieves number-free motion generation via P-Flow for mitigating computational overheads and S-Flow for alleviating error accumulation.
Extensive results show UMF achieves state-of-the-art performance for multi-person generation, and exhibits robust zero-shot generalization to challenging group scenarios. While the $1+N$ paradigm enhances generalization, UMF remains constrained to medium sized group interactions ($\approx 10$ agents) centered on a primary agent. Future will explore leveraging visual priors from large-scale video diffusion models to scale synthesis to dense crowd dynamics ($\approx 100$ agents).


%

\newpage


\section*{Acknowledgments}
This work was supported by the K-CSC funding.
The authors acknowledge the use of King’s CREATE HPC. Retrieved March 24, 2026, from \url{https://doi.org/10.18742/rnvf-m076}.


{
    \small
    \bibliographystyle{ieeenat_fullname}
    \bibliography{main}
}

\end{document}